%% file: main.tex
\documentclass[review]{cas-sc}

\usepackage[numbers,sort&compress]{natbib}

\input{defs}

\RenewDocumentCommand\printorcid{}{}

\makeatletter
\renewcommand\paragraph{\@startsection{paragraph}{4}{\parindent}%
    {10pt \@plus0.01pt \@minus0.01pt}%
    {-6pt}%
    {\ssssparaindent%
     \sssssectionfont\bfseries\raggedright}}
\makeatother
\usepackage{lineno}

\DTLloaddb{comp}{tables/comparison.csv}
\DTLloaddb{cyr}{tables/crossyear.csv}
\DTLloaddb{probe}{tables/probe.csv}
\DTLloaddb{src}{tables/source.csv}
\DTLloaddb{inp}{tables/inputs.csv}
\DTLloaddb{dsp}{tables/designspace.csv}
\DTLloaddb{prs}{tables/pairs.csv}
\DTLloaddb{lat}{tables/lattice.csv}
\DTLloaddb{rmp}{tables/ramps.csv}
\DTLloaddb{rsc}{tables/ramp_scores.csv}
\DTLloaddb{rgm}{tables/regime.csv}
\DTLloaddb{wrp}{tables/widthramp.csv}
\DTLloaddb{phz}{tables/perhorizon.csv}
\DTLloaddb{img}{tables/imagemetrics.csv}
\DTLloaddb{elv}{tables/elevation.csv}
\DTLloaddb{trn}{tables/training.csv}
\DTLloaddb{ctg}{tables/contiguity.csv}

\begin{document}
\begin{filecontents*}{tables/comparison.csv}
group,method,rmse,rmseci,fs,fsci,crps,crpsci,crpss,crpssci,brmse,bfs,bcrps,bcrpss
0,SPMref,127.4,"[88.7, 159.6]",--,,--,,--,,relax,relax,relax,relax
1,DLinearNoIclr,113.4,"[83.8, 138.4]",+0.110,"[+0.044, +0.144]",60.6,"[45.0, 76.6]",-0.081,"[-0.225, -0.004]",relax,relax,relax,relax
1,PatchTSTNoIclr,111.0,"[81.4, 135.6]",+0.128,"[+0.069, +0.163]",56.7,"[39.8, 73.9]",-0.011,"[-0.091, +0.036]",relax,relax,relax,relax
1,TimeXerNoIclr,111.0,"[81.6, 135.3]",+0.129,"[+0.064, +0.166]",57.4,"[40.9, 74.2]",-0.023,"[-0.113, +0.029]",relax,relax,relax,relax
2,DLinearIclr,109.8,"[80.7, 134.2]",+0.138,"[+0.079, +0.170]",58.0,"[42.2, 74.2]",-0.034,"[-0.156, +0.034]",relax,relax,relax,relax
2,PatchTSTIclr,109.6,"[80.0, 133.7]",+0.140,"[+0.078, +0.179]",54.5,"[37.4, 72.0]",+0.029,"[-0.022, +0.062]",relax,relax,relax,relax
2,TimeXerIclr,110.0,"[80.6, 134.0]",+0.137,"[+0.073, +0.177]",55.5,"[38.9, 72.4]",+0.011,"[-0.063, +0.054]",relax,relax,relax,relax
3,SunSet,97.6,"[72.7, 119.2]",+0.234,"[+0.171, +0.274]",52.7,"[38.5, 67.7]",+0.060,"[-0.061, +0.129]",bestone,bestone,relax,relax
4,ConvLSTM,145.3,"[104.6, 178.4]",-0.141,"[-0.272, -0.041]",77.0,"[56.2, 98.3]",-0.372,"[-0.562, -0.263]",relax,relax,relax,relax
4,PhyDNet,115.8,"[87.4, 139.2]",+0.091,"[-0.046, +0.180]",62.9,"[48.0, 78.0]",-0.121,"[-0.334, -0.002]",relax,relax,relax,relax
4,PhyDNetGAN,112.3,"[85.3, 134.8]",+0.118,"[-0.004, +0.197]",61.8,"[47.7, 76.1]",-0.103,"[-0.307, +0.011]",relax,relax,relax,relax
5,SkyGPT,216.8,"[178.0, 249.5]",-0.702,"[-1.093, -0.525]",121.1,"[99.6, 141.5]",-1.159,"[-1.864, -0.788]",relax,relax,relax,relax
5,Xiong,116.9,"[87.4, 141.0]",+0.082,"[-0.024, +0.146]",47.8,"[36.3, 59.6]",+0.148,"[+0.001, +0.231]",relax,relax,relax,relax
6,propFuture,104.1,"[78.3, 126.4]",+0.183,"[+0.098, +0.235]",36.9,"[27.1, 47.2]",+0.341,"[+0.250, +0.396]",besttwo,besttwo,bestone,bestone
6,propPast,114.1,"[84.7, 138.3]",+0.104,"[+0.013, +0.158]",42.9,"[31.0, 54.8]",+0.235,"[+0.140, +0.294]",relax,relax,besttwo,besttwo
\end{filecontents*}

\begin{filecontents*}{tables/crossyear.csv}
group,method,rin,rcy,rdel,fin,fcy,fdel,cin,ccy,cdel,brin,brcy,brdel,bfin,bfcy,bfdel,bcin,bccy,bcdel
0,SPMref,127.4,--,--,---,---,---,--,--,--,relax,relax,relax,relax,relax,relax,relax,relax,relax
1,PatchTSTIclr,109.6,109.0,$-0.6$,$+0.140$,$+0.145$,$+0.004$,+0.029,+0.031,$+0.002$,relax,relax,relax,relax,relax,relax,relax,relax,besttwo
1,SunSet,97.6,98.1,$+0.5$,$+0.234$,$+0.230$,$-0.004$,+0.060,+0.047,$-0.013$,bestone,bestone,relax,bestone,bestone,relax,relax,relax,relax
1,PhyDNetGAN,112.3,110.6,$-1.8$,$+0.118$,$+0.132$,$+0.014$,-0.103,-0.073,$+0.030$,relax,relax,relax,relax,relax,relax,relax,relax,relax
1,Xiong,116.9,117.2,$+0.3$,$+0.082$,$+0.080$,$-0.002$,+0.148,+0.148,$+0.000$,relax,relax,besttwo,relax,relax,besttwo,relax,relax,bestone
2,propPast,114.1,113.9,$-0.2$,$+0.104$,$+0.106$,$+0.001$,+0.235,+0.241,$+0.006$,relax,relax,bestone,relax,relax,bestone,besttwo,besttwo,relax
2,propFuture,104.1,105.7,$+1.7$,$+0.183$,$+0.170$,$-0.013$,+0.341,+0.343,$+0.002$,besttwo,besttwo,relax,besttwo,besttwo,relax,bestone,bestone,besttwo
\end{filecontents*}

\begin{filecontents*}{tables/probe.csv}
what,who,rmse,best
true future frames,frozen U-Net,63.5,relax
VAE reconstruction of the true frames,frozen U-Net,80.1,relax
last input frame repeated (persistence),frozen U-Net,127.9,relax
generated frames: one member,frozen U-Net,150.8,relax
generated frames: mean of ten members,frozen U-Net,132.2,relax
denoised estimate of the future sky: one member,irradiance head,134.6,relax
denoised estimate of the future sky: mean of ten members,irradiance head,104.1,bfmark
\end{filecontents*}

\begin{filecontents*}{tables/source.csv}
branch,arm,rmse,fs,crpss,ssr,picp,brmse,bfs,bcrpss,bssr,bpicp
none,past frames (scalar only),113.84,+0.106,+0.219,1.48,0.424,besttwo,besttwo,relax,relax,relax
yes,past frames (past sky),114.50,+0.101,+0.229,1.48,0.458,relax,relax,relax,relax,relax
yes,noised future frames (noised state),114.68,+0.100,+0.261,0.98,0.678,relax,relax,besttwo,bestone,bestone
yes,noised future and past frames (concat),113.96,+0.105,+0.246,1.41,0.498,relax,relax,relax,relax,relax
yes,denoised estimate of the future sky (future sky),104.58,+0.179,+0.334,1.18,0.615,bestone,bestone,bestone,besttwo,besttwo
yes,true future frames in training form (oracle; not a forecast),101.61,+0.202,+0.381,1.13,0.640,relax,relax,relax,relax,relax
\end{filecontents*}

\begin{filecontents*}{tables/inputs.csv}
seed,control,stream,gain,crpssgain
1,108.51,101.45,$+7.06$,$+0.090$
2,108.13,103.37,$+4.76$,$+0.084$
\end{filecontents*}

\begin{filecontents*}{tables/designspace.csv}
group,arm,rmse,fs,crpss,ssr,brmse,bfs,bcrpss,bssr
1,propFuture,104.58,+0.179,+0.334,1.18,bestone,bestone,bestone,bestone
1,ArmFiLMmatched,109.06,+0.144,+0.311,1.26,relax,relax,besttwo,relax
1,ArmFiLM,105.97,+0.168,+0.334,1.19,besttwo,besttwo,bestone,besttwo
2,propPast,114.50,+0.101,+0.229,1.48,relax,relax,relax,relax
2,ArmFiLMPast,114.26,+0.103,+0.233,1.46,relax,relax,relax,relax
\end{filecontents*}

\begin{filecontents*}{tables/pairs.csv}
without,withp,drmse,dcrpss,confounds
propFuture,ArmFiLMmatched,"$+4.5$ $[+0.7, +7.5]$","$-0.023$ $[-0.044, +0.009]$",starting checkpoint
propFuture,ArmFiLM,"$+1.4$ $[-0.4, +2.8]$","$+0.001$ $[-0.013, +0.024]$",starting checkpoint; resumed run
propPast,ArmFiLMPast,"$-0.2$ $[-0.4, -0.1]$","$+0.004$ $[+0.003, +0.006]$",selection rule; training budget
\end{filecontents*}

\begin{filecontents*}{tables/lattice.csv}
group,arm,scalar,image,loop,bflag
1,propPast,"$\vect{e},\vect{y}_t,\vect{a}$","$\vect{e},s_t,\vect{a}^i$",open,no
1,ArmNoisy,"$\vect{e},\vect{y}_t,\vect{x}_t$","$\vect{e},s_t,\vect{a}^i$",open,no
1,ArmConcat,"$\vect{e},\vect{y}_t,\vect{x}_t,\vect{a}$","$\vect{e},s_t,\vect{a}^i$",open,no
1,propFuture,"$\vect{e},\vect{y}_t,\hat{\vect{x}}_0$ ($\vect{a}$ ind.)","$\vect{e},s_t,\vect{a}^i$",open,no
1,ArmFiLMPast,"$\vect{e},\vect{y}_t,\vect{a}$","$\vect{e},s_t,\vect{a}^i,\hat{\vect{y}}_0$",open,no
1,ArmFiLM,"$\vect{e},\vect{y}_t,\hat{\vect{x}}_0$ ($\vect{a}$ ind.)","$\vect{e},s_t,\vect{a}^i,\hat{\vect{y}}_0$",open,no
1,ArmScalar,"$\vect{e},\vect{y}_t,\vect{a}$",--- (no image branch),n/a,no
1,ArmOracle,"$\vect{e},\vect{y}_t$, \textbf{true} $\vect{x}_0$","$\vect{e},s_t,\vect{a}^i$",open,no
2,ConvLSTM,frozen U-Net on gen. frames,$\vect{a}$,n/a,no
2,PhyDNet,frozen U-Net on gen. frames,$\vect{a}$,n/a,no
2,SkyGPT,frozen U-Net on gen. frames,$\vect{a}$,n/a,no
2,Xiong,"$\vect{b}$, gen. frames, temporal cov.",$\vect{a}$,n/a,yes
2,SunSet,"$\vect{b},\vect{a}$ (no generation)",---,n/a,yes
2,PVMM,"$\vect{a},\vect{b},\vect{y}_\tau,\vect{x}_\tau$","$\vect{a},\vect{b},\vect{y}_\tau,\vect{x}_\tau$",\textbf{closed},\textbf{yes}
\end{filecontents*}

\begin{filecontents*}{tables/ramps.csv}
method,picpn,picpr,picpd,ssn,ssr,ssd,bpicpn,bpicpr,bpicpd,bssn,bssr,bssd
DetGroup,0.000,0.000,0,---,---,---,relax,relax,relax,relax,relax,relax
SkyGPT,0.235,0.236,$-0.001$,3.076,3.353,$+0.277$,relax,relax,bestone,relax,relax,relax
Xiong,0.847,0.642,$-0.206$,0.806,1.796,$+0.990$,bestone,bestone,relax,relax,relax,relax
propPast,0.458,0.518,$+0.060$,1.282,1.515,$+0.233$,relax,relax,relax,besttwo,besttwo,bestone
propFuture,0.631,0.605,$-0.026$,0.972,1.227,$+0.255$,besttwo,besttwo,besttwo,bestone,bestone,besttwo
\end{filecontents*}

\begin{filecontents*}{tables/ramp_scores.csv}
group,method,rmse,fs,crpss,brmse,bfs,bcrpss
1,DLinearNoIclr,249.8,$+0.147$,$+0.109$,relax,relax,relax
1,PatchTSTNoIclr,244.8,$+0.164$,$+0.111$,relax,relax,relax
1,TimeXerNoIclr,244.3,$+0.166$,$+0.112$,relax,relax,relax
2,DLinearIclr,242.3,$+0.173$,$+0.125$,relax,relax,relax
2,PatchTSTIclr,243.3,$+0.170$,$+0.113$,relax,relax,relax
2,TimeXerIclr,243.3,$+0.169$,$+0.115$,relax,relax,relax
3,SunSet,216.4,$+0.261$,$+0.198$,bestone,bestone,relax
4,ConvLSTM,324.2,$-0.107$,$-0.159$,relax,relax,relax
4,PhyDNet,254.3,$+0.132$,$+0.109$,relax,relax,relax
4,PhyDNetGAN,245.1,$+0.163$,$+0.137$,relax,relax,relax
5,SkyGPT,406.5,$-0.388$,$-0.334$,relax,relax,relax
5,Xiong,254.0,$+0.133$,$+0.310$,relax,relax,relax
6,propFuture,226.4,$+0.227$,$+0.440$,besttwo,besttwo,bestone
6,propPast,249.0,$+0.150$,$+0.342$,relax,relax,besttwo
\end{filecontents*}

\begin{filecontents*}{tables/regime.csv}
tertile,cprop,csun,clead,rprop,rsun,bcprop,bcsun,bclead,brprop,brsun
Overcast,$+0.285$,$-0.002$,$+0.287$,101.6,92.6,bfmark,relax,relax,relax,bfmark
Variable,$+0.347$,$+0.085$,$+0.262$,104.9,98.5,bfmark,relax,relax,relax,bfmark
Clear,$+0.388$,$+0.091$,$+0.297$,105.6,101.5,bfmark,relax,relax,relax,bfmark
\end{filecontents*}

\begin{filecontents*}{tables/widthramp.csv}
method,r,rho,wnon,wramp,ratio,br,brh
SkyGPT,$+0.6574$,$+0.5941$,85.4,266.0,3.1,relax,relax
Xiong,$+0.6458$,$+0.6285$,152.6,346.2,2.3,relax,relax
propPast,$+0.8034$,$+0.8260$,61.7,355.3,5.8,besttwo,bestone
propFuture,$+0.8457$,$+0.8218$,87.1,417.1,4.8,bestone,besttwo
\end{filecontents*}

\begin{filecontents*}{tables/perhorizon.csv}
h,rgt,rgen,crpss,p90,dfs,dfssun
i+1,65.7,124.7,$+0.034$,0.660,$+0.015$,$+0.004$
i+2,61.3,123.9,$+0.252$,0.615,$+0.018$,$+0.006$
i+3,63.8,127.5,$+0.325$,0.634,$+0.022$,$+0.005$
i+4,63.1,128.4,$+0.375$,0.600,$-0.027$,$-0.001$
i+5,65.3,135.4,$+0.410$,0.623,$-0.008$,$+0.004$
i+6,61.8,136.5,$+0.397$,0.613,$-0.012$,$-0.003$
i+7,62.5,136.1,$+0.398$,0.662,$-0.038$,$-0.010$
i+8,64.0,144.1,$+0.385$,0.606,$-0.053$,$-0.030$
\end{filecontents*}

\begin{filecontents*}{tables/imagemetrics.csv}
grp,arm,frames,lpips,psnr,rmse,bl,bp,br
1,RefVAE,---,0.0056,31.89,80.1,relax,relax,relax
1,RefPersist,---,0.0427,19.65,127.9,bestone,relax,besttwo
2,ConvLSTM,single,0.0845,21.99,145.3,relax,besttwo,relax
2,PhyDNet,single,0.1025,22.59,115.8,relax,besttwo,besttwo
2,PhyDNetGAN,single,0.1028,22.63,112.3,relax,bestone,bestone
3,SkyGPT,one member,0.0781,19.15,226.1,relax,relax,relax
3,SkyGPT,mean of ten,0.0963,20.34,216.9,relax,relax,relax
3,Xiong,one member,0.0601,19.74,185.2,relax,relax,relax
3,Xiong,mean of ten,0.0846,21.29,171.2,relax,relax,relax
3,propPast,one member,0.0591,20.68,161.7,besttwo,relax,relax
3,propPast,mean of ten,0.0826,21.76,146.6,relax,relax,relax
3,propFuture,one member,0.0577,20.76,150.8,besttwo,relax,relax
3,propFuture,mean of ten,0.0815,21.77,132.2,relax,relax,relax
\end{filecontents*}

\begin{filecontents*}{tables/elevation.csv}
arm,r0,c0,r10,c10,r20,c20,br0,bc0,br10,bc10,br20,bc20
SPMrow,127.4,---,140.4,---,165.3,---,relax,relax,relax,relax,relax,relax
propFuture,104.1,+0.341,114.6,+0.347,134.3,+0.365,besttwo,bestone,besttwo,bestone,besttwo,bestone
propPast,114.1,+0.235,125.7,+0.238,147.5,+0.257,relax,besttwo,relax,besttwo,relax,besttwo
Xiong,116.9,+0.148,128.8,+0.148,151.1,+0.161,relax,relax,relax,relax,relax,relax
SunSet,97.6,+0.060,107.5,+0.062,125.9,+0.083,bestone,relax,bestone,relax,bestone,relax
PhyDNetGAN,112.3,$-0.103$,123.4,$-0.080$,143.9,$-0.032$,relax,relax,relax,relax,relax,relax
\end{filecontents*}

\begin{filecontents*}{tables/training.csv}
stage,trains,frozen,lr,stop
1,VAE,---,$10^{-6}$,early stopping on validation LPIPS (patience 20)
2,UNet,VAE,"$10^{-5}$, then $10^{-6}$ for the final epochs",best validation LPIPS (patience 20)
3,ControlNet,"VAE, UNet","$10^{-5}$, cosine schedule",best validation LPIPS (patience 20)
4,"Joint diffusion training (phase 1): image tokenizer, irradiance head, temporal refiner","VAE, UNet, ControlNet",$10^{-4}$,19--20 epochs; best validation LPIPS $+$ CSI RMSE
5,Joint diffusion training (phase 2): all except the VAE,VAE,"$10^{-6}$, halved on plateau",at most 31 epochs; best validation LPIPS $+$ CSI RMSE
\end{filecontents*}

\begin{filecontents*}{tables/contiguity.csv}
group,method,rall,rex,call,cex,rqc,cqc,brall,brex,brqc,bcall,bcex,bcqc
0,SPMref,127.4,127.8,--,--,128.7,--,relax,relax,relax,relax,relax,relax
1,DLinearNoIclr,113.4,113.8,$-0.081$,$-0.081$,114.2,$-0.074$,relax,relax,relax,relax,relax,relax
1,PatchTSTNoIclr,111.0,111.4,$-0.011$,$-0.010$,111.9,$-0.008$,relax,relax,relax,relax,relax,relax
1,TimeXerNoIclr,111.0,111.4,$-0.023$,$-0.023$,111.8,$-0.019$,relax,relax,relax,relax,relax,relax
2,DLinearIclr,109.8,110.1,$-0.034$,$-0.035$,110.6,$-0.033$,relax,relax,relax,relax,relax,relax
2,PatchTSTIclr,109.6,109.9,$+0.029$,$+0.029$,110.3,$+0.031$,relax,relax,relax,relax,relax,relax
2,TimeXerIclr,110.0,110.3,$+0.011$,$+0.011$,110.7,$+0.013$,relax,relax,relax,relax,relax,relax
3,SunSet,97.6,97.9,$+0.060$,$+0.060$,98.5,$+0.059$,bestone,bestone,bestone,relax,relax,relax
4,ConvLSTM,145.3,145.7,$-0.372$,$-0.372$,146.5,$-0.374$,relax,relax,relax,relax,relax,relax
4,PhyDNet,115.8,116.1,$-0.121$,$-0.120$,117.0,$-0.124$,relax,relax,relax,relax,relax,relax
4,PhyDNetGAN,112.3,112.7,$-0.103$,$-0.102$,113.5,$-0.106$,relax,relax,relax,relax,relax,relax
5,SkyGPT,216.8,217.5,$-1.159$,$-1.160$,219.1,$-1.169$,relax,relax,relax,relax,relax,relax
5,Xiong,116.9,117.3,$+0.148$,$+0.148$,118.0,$+0.145$,relax,relax,relax,relax,relax,relax
6,propFuture,104.1,104.4,$+0.341$,$+0.341$,104.8,$+0.342$,besttwo,besttwo,besttwo,bestone,bestone,bestone
6,propPast,114.1,114.5,$+0.235$,$+0.235$,115.1,$+0.234$,relax,relax,relax,besttwo,besttwo,besttwo
\end{filecontents*}

\let\WriteBookmarks\relax
\def\floatpagepagefraction{1}
\def\textpagefraction{.001}
\shorttitle{Single-stage joint diffusion for irradiance forecasting}
\shortauthors{K. Suwandee et~al.}

\title [mode = title]{Single-stage joint diffusion of sky images and clear-sky index for
ultra-short-term irradiance forecasting: what the frame-to-irradiance stage discards}

\author[1]{Kanawut Suwandee}
\author[1]{Suwichaya Suwanwimolkul}
\cormark[1]
\ead{suwichaya.s@chula.ac.th}

\affiliation[1]{organization=Chulalongkorn University,  
            addressline={254 Phaya Thai Rd, Wang Mai, Pathum Wan}, 
            city=Bangkok,
            postcode={10330},  
            country=Thailand}

\cortext[1]{Corresponding author}

\begin{abstract}
Sky-image irradiance forecasting is dominated by a two-stage design: a video-prediction model
forecasts future sky frames, and a separately trained regressor maps those frames to irradiance. We
propose a single-stage alternative that co-denoises the future sky-image and clear-sky-index
sequences in one diffusion process, with an irradiance head that reads a denoised estimate of the
future sky rather than the noised diffusion state. The forecaster needs no regression stage and
runs from sky images and deterministic covariates alone, with no measured irradiance
at its input, the situation at a site with a camera but no pyranometer. On 2023 SIRTA data the
model reaches a pooled RMSE of $104.1$~W\,m$^{-2}$ and a skill score of $+0.34$ on the continuous
ranked probability score (CRPS) against smart persistence, and retains that skill when retrained on
2021--2022. Reading the denoised estimate, the model is
$8.3$~W\,m$^{-2}$ more accurate than the best two-stage cascade on the same samples; reading the
model's own frames through a separately trained regressor instead costs $28$~W\,m$^{-2}$. Among the
generative forecasters its frames are also the closest to the observed sky. Point accuracy is
competitive but not best: a deterministic network that also reads measured irradiance attains
$97.6$~W\,m$^{-2}$, yet the model's CRPS and CRPS skill score are better than those of all twelve
baseline configurations. Code is available at
\url{https://github.com/GenAI-CUEE/single-stage-joint-diffusion}.
\end{abstract}

\begin{highlights}
\item Forecasts irradiance from sky images alone, with no measured irradiance at its input.
\item Removes the frame-to-irradiance regression stage of the two-stage cascade.
\item Reaches $+0.34$ CRPS skill over smart persistence, ahead of twelve baselines.
\item Its prediction-interval width tracks the size of the coming ramp ($r=0.85$).
\item Ablation: the denoised future-sky estimate beats past frames and the noised state.
\end{highlights}


\maketitle

\clearpage 

\input{sec_intro}

\input{sec_litreview}

\input{sec_method}

\input{sec_setup}

\input{sec_comparison}

\input{sec_cascade}

\input{sec_imagequality}

\input{sec_calibration}

\input{sec_ablations}

\input{sec_conclusion}

\input{sec_appendix}

\section*{CRediT authorship contribution statement}
\textbf{Kanawut Suwandee:} Software, Validation, Formal analysis, Investigation, Data curation,
Visualization, Writing --- original draft.
\textbf{Suwichaya Suwanwimolkul:} Conceptualization, Methodology, Supervision, Project
administration, Funding acquisition, Writing --- review and editing.

\section*{Declaration of competing interest}
The authors declare that they have no known competing financial interests or personal relationships
that could have appeared to influence the work reported in this paper.

\section*{Funding}
This work was supported by the 111\textsuperscript{th} Anniversary Engineering Research Catalyst
Fund Towards U Top 100, Chulalongkorn University.

\section*{Data availability}
The code, the trained weights and stored forecasts of every method compared here, and the training
scripts are available at \url{https://github.com/GenAI-CUEE/single-stage-joint-diffusion}. Scoring
the stored forecasts reproduces every entry of Table~\ref{tab:comparison} exactly. On the data as
used here, regenerating the forecasts from the released weights reproduces the deterministic methods and
the proposed model, whose sampling is seeded, exactly. \SkyGPT{}'s sampler
is unseeded and the forward pass of \citet{xiong2025multimodal} is not bit-for-bit deterministic, so
their ensembles change from run to run; regenerated scores fall within the confidence intervals
reported in Table~\ref{tab:comparison}. Reproducing the proposed model's own training additionally requires the stage schedule of
Appendix~\ref{sec:app-protocol}, which the released configurations do not set.

The sky images and irradiance measurements are provided by the SIRTA atmospheric
observatory~\citep{haeffelin2005sirta} under its own data policy and are not redistributed here. The
repository includes scripts that retrieve them from SIRTA's server, to which SIRTA grants access on
request, and the preprocessing code that builds the dataset; it also lists the identifiers and frame
timestamps of the samples used. Because of SIRTA's review of the 2023 record (Section~\ref{sec:data}),
a dataset built from today's files lacks some of the test samples and builds others from shifted
frames. The repository lists the samples such a rebuild lacks, provides a script that identifies the
shifted ones from the frame timestamps, and gives the expected scores for the rebuild.

\section*{Declaration of generative AI and AI-assisted technologies in the manuscript preparation process}
During the preparation of this work the authors used Claude (Anthropic) in order to
draft and revise text, check the consistency of the manuscript with the code, and prepare
analysis scripts. After using this tool, the authors reviewed and edited the content as needed and
take full responsibility for the content of the published article.

\section*{Acknowledgements}
The authors would like to acknowledge SIRTA for providing the data used in this
study~\citep{haeffelin2005sirta}.

\bibliographystyle{elsarticle-num-names}

\bibliography{references}

\end{document}

%% file: defs.tex
\usepackage{caption}
\usepackage{subcaption}
\usepackage{amsmath}
\usepackage{amssymb}
\usepackage{amsfonts}
\usepackage{bm}

\newcommand{\eg}{{\it e.g.}}

\newcommand{\vect}[1]{\bm{#1}}

\newcommand{\Set}[1]{\{ #1 \}}
\newcommand{\Brc}[1]{\left( #1 \right) }

\newcommand{\propFuture}{\textit{Proposed} (joint, future sky)}
\newcommand{\propPast}{\textit{Proposed} (joint, past sky)}

\newcommand{\ArmNoisy}{\textit{noised state}}
\newcommand{\ArmConcat}{\textit{concat}}
\newcommand{\ArmFiLM}{\textit{Proposed} (joint, future sky, future CSI)}
\newcommand{\ArmFiLMPast}{\textit{Proposed} (joint, past sky, future CSI)}
\newcommand{\ArmFiLMmatched}{\textit{Proposed} (joint, future sky, future CSI), matched training budget}
\newcommand{\ArmScalar}{\textit{scalar only}}

\newcommand{\PVMM}{PV-MM-Diffusion}

\newcommand{\Xiong}{Xiong}
\newcommand{\SkyGPT}{SkyGPT}
\newcommand{\SunSet}{SUNSET}
\newcommand{\ConvLSTM}{ConvLSTM}
\newcommand{\PhyDNet}{PhyDNet}
\newcommand{\PhyDNetGAN}{PhyDNetGAN}
\newcommand{\DLinearIclr}{DLinear}
\newcommand{\PatchTSTIclr}{PatchTST}
\newcommand{\TimeXerIclr}{TimeXer}

\usepackage{array}
\newcolumntype{P}[1]{>{\hspace{0pt}\centering\arraybackslash}p{#1}}   
\newcolumntype{Q}[1]{>{\hspace{0pt}\raggedright\arraybackslash}p{#1}} 
\newcolumntype{S}[1]{>{\hspace{0pt}\raggedleft\arraybackslash}p{#1}}  

\newcommand{\wMethod}{0.235\textwidth}   
\newcommand{\wRMSE}  {0.160\textwidth}   
\newcommand{\wFS}    {0.205\textwidth}   
\newcommand{\wCRPS}  {0.155\textwidth}   
\newcommand{\wCRPSS} {0.205\textwidth}   
\newcommand{\wProbeWhat} {0.560\linewidth}  
\newcommand{\wProbeWho}  {0.200\linewidth}  
\newcommand{\wProbeRMSE} {0.220\linewidth}  
\newcommand{\wSrcBr}    {0.090\linewidth}  
\newcommand{\wSrcArm}   {0.300\linewidth}  
\newcommand{\wSrcN}     {0.112\linewidth}  
\newcommand{\wInpSeed}  {0.120\linewidth}  
\newcommand{\wInpCtl}   {0.180\linewidth}
\newcommand{\wInpStr}   {0.220\linewidth}
\newcommand{\wInpGain}  {0.220\linewidth}
\newcommand{\wInpCrpss} {0.220\linewidth}

\newcommand{\wPhzH}  {0.090\linewidth}   
\newcommand{\wPhzN}  {0.135\linewidth}
\newcommand{\wCtgM}  {0.280\linewidth}   
\newcommand{\wCtgN}  {0.105\linewidth}
\newcommand{\wWrpM}  {0.290\linewidth}   
\newcommand{\wWrpN}  {0.125\linewidth}
\newcommand{\wRgmT}  {0.150\linewidth}   
\newcommand{\wRgmN}  {0.155\linewidth}
\newcommand{\wRmpMethod}{0.300\linewidth}  
\newcommand{\wRmpNum}   {0.200\linewidth}
\newcommand{\wLatArm}{0.350\textwidth}  
\newcommand{\wLatSc} {0.250\textwidth}
\newcommand{\wLatIm} {0.230\textwidth}
\newcommand{\wLatL}  {0.065\textwidth}
\newcommand{\wLatB}  {0.055\textwidth}
\newcommand{\wDsArm}{0.540\linewidth}  
\newcommand{\wDsN}  {0.105\linewidth}

\newcommand{\wRmpM}    {0.280\linewidth}  
\newcommand{\wRmpC}    {0.105\linewidth}
\newcommand{\wRmpD}    {0.145\linewidth}
\newcommand{\wMetricCYn}{0.083\linewidth}  
\newcommand{\wMethodCY}{0.230\linewidth} 
\newcommand{\wImgM}{0.340\linewidth}  
\newcommand{\wImgK}{0.160\linewidth}  
\newcommand{\wImgN}{0.160\linewidth}  
\newcommand{\wElvM}{0.250\linewidth}  
\newcommand{\wElvN}{0.120\linewidth}  

\newcommand{\arrowdown}{$\downarrow$}
\newcommand{\arrowup}{$\uparrow$}
\newcommand{\arrowzero}{$\rightarrow\!0$}

\usepackage[normalem]{ulem}  

\newcommand{\wTrSt}{0.055\linewidth}
\newcommand{\wTrTr}{0.270\linewidth}
\newcommand{\wTrFr}{0.180\linewidth}
\newcommand{\wTrLr}{0.160\linewidth}
\newcommand{\wTrSp}{0.300\linewidth}

\usepackage{tikz}
\usetikzlibrary{positioning,fit,arrows.meta,calc}

\usepackage{datatool}

%% file: sec_intro.tex
\section{Introduction}
\label{sec:intro}

Global installed photovoltaic capacity passed $2.2$~TW in 2024~\citep{ieapvps2025snapshot}, and its output varies on
timescales set by cloud motion. Energy storage, such as batteries, can absorb part of this variability, at its own cost; the rest
must be met by the remainder of the system, which schedules against
forecasts~\citep{otuka2025experimental}. The hardest horizons are minutes to an hour ahead, where
cloud-driven ramps coincide with large net-load ramps~\citep{varaschin2025benchmarking} and numerical
weather prediction, designed for longer lead times, contributes little.

Ground-based all-sky imagers observe the clouds about to pass between the sun and the array, and a
large body of work maps sky images to irradiance, from cloud-motion tracking to convolutional and
recurrent networks~\citep{paletta2021benchmarking}. Mapping an image to its \emph{concurrent}
irradiance (estimation) is, however, much easier than forecasting irradiance minutes \emph{ahead}
(nowcasting): \citet{varaschin2025benchmarking} report estimation errors far below what any forecast at
the same site achieves.

Beyond point accuracy, the uncertainty of the prediction also matters. One group of works moves from deterministic point forecasts to
probabilistic ones --- what an operator planning reserve actually needs --- which brings with it 
questions about which scores are appropriate, what a fair reference forecast is, and how calibration
should be verified~\citep{gneiting2023probabilistic}. Another moves from regressing irradiance directly out of observed frames to \emph{generating} the future sky and forecasting from
it. SkyGPT~\citep{nie2024skygpt} is the canonical instance, pairing a stochastic video predictor with
a separately trained irradiance regressor. Two works concurrent with this one also generate the
future sky: PV-MM-Diffusion~\citep{huang2026pvmm} forecasts sky images and photovoltaic power jointly
in a single multimodal diffusion process, and \citet{xiong2025multimodal} couple a neural stochastic
differential equation video model to a probabilistic forecasting head. The video-prediction lineage those methods draw
on --- ConvLSTM~\citep{shi2015convlstm} and PhyDNet~\citep{leguen2020phydnet} --- supplies the
deterministic rollout baselines we reimplement here.

 The dominant paradigm is \emph{two-stage}: a
video-prediction model forecasts the future sky, and a separate regressor maps each generated frame
to an irradiance value. The design is intuitive --- forecast the sky, then read the irradiance off it; however, the forecast sky is a high-dimensional image sequence, an intermediate to predicting a scalar, and
the irradiance estimate inherits \emph{both} the frame-generation error and the frame-to-irradiance
regression error. Frame generation at these horizons is hard and imperfect for every method we
evaluate; in a two-stage design that imperfection propagates directly into the forecast.

We instead co-denoise the future frames and the future clear-sky-index sequence in a \emph{single}
diffusion process with an \emph{image branch} and an \emph{irradiance head}, and read the forecast directly from
the irradiance head. A diffusion model generates by removing noise step by step. At every step,
although the image branch's state is still noisy, the model can compute a clean estimate of the
finished future frames: a \emph{denoised estimate} of the future sky. The irradiance head reads this
estimate, not the noisy state, together with the clock variables and the clear-sky irradiance at the
forecast origin, to provide the irradiance forecast. The model receives no measured irradiance
history, so it can serve a site instrumented with a camera but no pyranometer. Providing the model's
\emph{own} generated frames to a separately trained regressor gives $132.2$~W\,m$^{-2}$, whereas the
jointly trained irradiance head reading the same generated sky reaches $104.1$. The irradiance head is
also more accurate than the best two-stage cascade we evaluate.

\paragraph{Contributions.} The proposed design departs from existing sky-image forecasters in two
respects. Unlike two-stage methods such as \SkyGPT{}~\citep{nie2024skygpt} and the model of
\citet{xiong2025multimodal}, it requires no separately trained frame-to-irradiance regressor. Unlike the
concurrent \PVMM{}~\citep{huang2026pvmm}, its irradiance head conditions on a denoised estimate of the
future sky rather than on the noised diffusion state (Appendix~\ref{sec:lattice} lists what each method
conditions on). Beyond proposing the single-stage design, we also ablate which
information the co-denoising should pass to the irradiance head. In every arm, the irradiance head receives the same clock variables and clear-sky irradiance. The arms
differ in which frames it reads and in whether an image branch is trained with it. In \ArmScalar{}, the
irradiance head reads the past frames and there is no image branch. In \textit{past sky}, the
irradiance head reads the past frames and is trained jointly with an image branch that generates the
future frames. In \ArmNoisy{}, the irradiance head reads the noised future frames of the image branch at
each denoising step, and in \ArmConcat{} it reads the past frames and the noised state together (the
information set of \PVMM{}). In \textit{future sky}, the proposed model, the irradiance head reads the
denoised estimate of the future sky. A sixth arm, \textit{oracle}, gives the trained irradiance head of
\textit{future sky} the true future frames, as an upper bound. Apart from the oracle, only
\textit{future sky} improves over \ArmScalar{} in RMSE
(Section~\ref{sec:ablation-source}). The model is built on a
pretrained Stable Diffusion~1.5 backbone with ControlNet, extended with a temporal refiner and an
irradiance head, rather than on a task-specific architecture. Our contributions are as follows.

\begin{enumerate}
\item \textbf{Irradiance head reading the denoised estimate.} Within each reverse diffusion step, the model forms a
denoised estimate of the future sky frames and conditions the irradiance head on it, removing the
frame-to-irradiance regression stage. Reading the denoised estimate of the future sky lowers RMSE by $9.3$~W\,m$^{-2}$ against
\ArmScalar{}, whereas \textit{past sky} gives no gain; given the true future frames (\textit{oracle}),
the trained irradiance head improves by only a further $3.0$~W\,m$^{-2}$.
The irradiance head reading the denoised estimate is also $8.3$~W\,m$^{-2}$ more accurate than the best two-stage
cascade on the same test samples, and $28$~W\,m$^{-2}$ more accurate than a separately trained U-Net
reading the model's own frames. Feeding the denoised estimate of the future clear-sky index back into the image branch brings no gain, and costs $4.5$~W\,m$^{-2}$ when trained to the
same budget.

\item \textbf{Competitive skill and sky-image quality.} Without access to measured irradiance, the
model attains a continuous ranked probability score (CRPS) skill score, CRPSS, of $+0.34$ relative to smart persistence and outperforms the model of
\citet{xiong2025multimodal}, which conditions on the irradiance history, in RMSE, forecast skill and
CRPSS. \SunSet{}~\citep{sun2019sunset}, a deterministic sky-image baseline that also reads the
irradiance history, retains lower pooled RMSE ($97.6$ against $104.1$~W\,m$^{-2}$) while providing
no predictive distribution. The frames the proposed model generates are also the closest to
the true sky among the generative forecasters, on both LPIPS ($0.058$ for a single member, against
$0.060$ for \Xiong{} and $0.078$ for \SkyGPT{}) and PSNR ($20.76$~dB, against $19.74$~dB and
$19.15$~dB); the deterministic video predictors reach a higher PSNR, but with blurred frames (Section~\ref{sec:imagequality}).

\item \textbf{A general-purpose, extensible framework.} Because the model is assembled from standard
components, its inputs can be varied directly. Adding the measured irradiance history as a single
additional input stream reduces RMSE by $4.8$--$7.1$~W\,m$^{-2}$ against a matched control, whereas a
perfect future sky is worth about $3$~W\,m$^{-2}$ to the trained irradiance head --- so the remaining
point-accuracy gap is better addressed through the inputs than through frame quality.

\item \textbf{Extensive and leak-free evaluation.} The model is compared against twelve baseline
configurations spanning scalar time-series, deterministic sky-image, video-prediction and generative
forecasters. All are scored on a shared test subset against a common smart-persistence reference,
with day-block bootstrap intervals and measured noise floors. A separate analysis of ramp events is
also provided. Across lead
times, the model's CRPSS is the highest of all methods at every one of the eight horizons, rising from
$+0.03$ at the first step to about $+0.4$, while its forecast skill stays below that of \SunSet{} at
every horizon (Section~\ref{sec:horizon}). Broken down by the sky condition at the forecast origin,
the model's CRPSS leads that of \SunSet{} in every tertile, by $0.26$ to $0.30$
(Section~\ref{sec:regimes}). To
exclude leakage through the few test days shared with training and validation, six methods, including the proposed model, are
additionally retrained on 2021--2022 and evaluated on the same 2023 test set: the model's CRPSS
is unchanged ($+0.343$ against $+0.341$), and its RMSE rises by $1.7$~W\,m$^{-2}$. Removing the
windows with low sun (solar elevation below $10^\circ$ or $20^\circ$ at the forecast origin), where the
clear-sky model is least reliable, leaves every paired comparison with the same sign
(Appendix~\ref{sec:app-protocol}).
\end{enumerate}

Section~\ref{sec:method} describes the model. Section~\ref{sec:setup} sets out the data and the scoring
conventions. Section~\ref{sec:comparison} reports the performance comparison;
Section~\ref{sec:cascade} quantifies the effect of removing the regression stage, and
Section~\ref{sec:imagequality} scores the generated frames directly; Section~\ref{sec:calibration} reports the per-horizon, sky-condition, uncertainty and
ramp analyses; and Section~\ref{sec:ablations} the two ablations.

%% file: sec_litreview.tex
\section{Literature review}
\label{sec:litreview}

\subsection{Ultra-short-term forecasting and its inputs}
\label{sec:lit-inputs}

At horizons of minutes to half an hour, irradiance variability is driven by individual clouds passing
between the sun and the sensor, a process that numerical weather prediction, and at the minute scale
also satellite products~\citep{aicardi2022satellite}, resolve too coarsely. Ground-based all-sky
imagers observe these clouds directly and have become the standard input at this
horizon~\citep{paletta2023advances}; \citet{nie2024datasets} survey the public datasets. The two used
most often are SKIPP'D~\citep{nie2023skippd}, which pairs images with the output of a $30$~kW rooftop
array, and the SIRTA records used here~\citep{haeffelin2005sirta}, which pair them with measured
global horizontal irradiance (GHI). Much recent work forecasts PV power at a single plant; this study
forecasts GHI, which does not depend on the response of a particular array.

\subsection{From sky images to irradiance: estimation and the two-stage cascade}
\label{sec:lit-regressors}
\label{sec:lit-cascade}

Mapping an observed sky image to the concurrent irradiance is a mature regression problem.
\citet{sun2019sunset} established the convolutional baseline that most later work reuses, and
\citet{paletta2021benchmarking} benchmarked the main architectural choices. Later work extended it to
image sequences~\citep{feng2022cnnintrahour} and to transformer encoders~\citep{liu2023multimodal};
combining sky images with real-time irradiance measurements still yields the strongest point forecasts
at this horizon~\citep{caldas2019veryshort}. \citet{varaschin2025benchmarking} quantify how well such
an estimator performs on real images, the natural reference for the estimation step of a two-stage
forecaster.

Forecasting requires the future sky, and the dominant design predicts future frames with a video model
and then applies such a regressor to them. The video stage has been built from convolutional recurrent
networks~\citep{shi2015convlstm}, physically disentangled dynamics~\citep{leguen2020phydnet},
adversarial and stochastic models~\citep{lee2018savp}, a purpose-built perturbation
model~\citep{paletta2022eclipse} and deep generative models~\citep{carpentieri2025generative}. Because
deterministic predictors blur as the horizon grows, \citet{nie2024skygpt} moved to stochastic
prediction, coupling a physics-constrained VideoGPT~\citep{yan2021videogpt} to a separately trained
U-Net regressor. In such a cascade the regressor is trained on real frames and never sees a generated one. The cost of
imperfect frames is usually bounded by applying the regressor to the \emph{real} future
image~\citep{nie2024skygpt,huang2026pvmm}; Section~\ref{sec:cascade} applies one regressor to real and
generated frames on the same samples.

\subsection{Probabilistic forecasting and its evaluation}
\label{sec:lit-probabilistic}

The verification framework for probabilistic forecasts is well
established~\citep{gneiting2014probabilistic,gneiting2023probabilistic}: proper scores such as the
CRPS, the coverage of prediction intervals, and the relation between ensemble spread and error. Solar
forecast distributions have been obtained by gradient
boosting~\citep{mitrentsis2022interpretable,song2025ngboost}, by Gaussian-process and transformer
hybrids~\citep{xiong2025transformer}, and by ensembles of generated futures~\citep{nie2024skygpt}.
Calibration diagnostics are reported far less often than point errors, and skill scores are quoted
against references that differ between papers. This study therefore fixes one reference, smart
persistence, and reports the CRPS skill score, interval coverage and the spread--skill ratio alongside
the point error (Section~\ref{sec:metrics}).

\subsection{Diffusion models and single-stage joint forecasting}
\label{sec:lit-diffusion}

Denoising diffusion models~\citep{ho2020ddpm} generate by reversing a noising process; the latent
formulation~\citep{rombach2022ldm} made high-resolution synthesis practical,
ControlNet~\citep{zhang2023controlnet} adds spatial conditioning to a pretrained backbone, and
FiLM~\citep{perez2018film} conditions intermediate features on an auxiliary signal. The proposed model
is assembled from these components. Two concurrent works generate the future sky alongside the forecast target, both for PV power at a
single site. \citet{huang2026pvmm} denoise sky images and PV power in two branches coupled by
cross-modal attention, and \citet{xiong2025multimodal} couple a stochastic video model to a
probabilistic head that also reads the measured power history. The
proposed irradiance head differs from both in what it reads: a denoised estimate of the future sky
formed inside each reverse step, rather than the noised state of the image branch (\PVMM{}) or frames
from a separate video stage (\Xiong{}).

%% file: sec_method.tex
\section{Method}
\label{sec:method}

\subsection{Preliminaries: denoising diffusion}
\label{sec:method-prelim}

A denoising diffusion model~\citep{ho2020ddpm} corrupts a data sample $u_0$ into noise over $T$ steps
by a fixed \emph{forward} process. With a variance schedule $\{\beta_t\}_{t=1}^{T}$, $\alpha_t=1-\beta_t$
and $\bar\alpha_t=\prod_{s\le t}\alpha_s$, the state at any step is available in closed form,
\begin{equation}
u_t = \sqrt{\bar\alpha_t}\,u_0 + \sqrt{1-\bar\alpha_t}\,\epsilon,
\qquad \epsilon\sim\mathcal{N}(0,\mathbf{I}).
\label{eq:forward}
\end{equation}
A network $\epsilon_\theta$ is trained to regress the added noise (the $\epsilon$-prediction
parameterization, used here),
\begin{equation}
\mathcal{L} = \mathbb{E}_{u_0,\epsilon,t}\big\|\epsilon - \epsilon_\theta(u_t,t,c)\big\|_2^2,
\label{eq:eps}
\end{equation}
with $c$ any conditioning information, and sampling applies the learned reverse step from $u_T$ to
$u_0$. Rearranging Eq.~\eqref{eq:forward} gives, at any step, the clean sample implied by the current
noise prediction,
\begin{equation}
\hat{u}_0 = \frac{u_t - \sqrt{1-\bar\alpha_t}\;\epsilon_\theta(u_t,t,c)}{\sqrt{\bar\alpha_t}}.
\label{eq:x0hat}
\end{equation}
Two objects therefore describe the same sample at every step: the \emph{noised iterate} $u_t$, which
the sampler advances, and the \emph{denoised estimate} $\hat{u}_0$, the model's current best guess at
the clean sample. The first is noisy by construction, increasingly so at large $t$; the second
estimates the noise-free target. Which of the two a downstream component reads is a design choice,
and Section~\ref{sec:ablation-source} shows that it matters.

\subsection{Problem setup}
\label{sec:method-setup}

At forecast origin $i$, the last observed time step, the model receives the eight preceding RGB sky
frames at two-minute cadence, $\vect{a}=\Set{\vect{a}^{j}}_{j=i-7}^{i}\in\mathbb{R}^{8\times3\times64\times64}$,
and the covariates of the most recent frame,
$\vect{e}=[\,\text{hour},\ \text{minute},\ \text{day of year},\ I_{clr}\,]\in\mathbb{R}^{4}$. The model
$F_\Theta(\cdot)$ is a conditional generator that jointly denoises, from $t=T$ to $t=0$, the latents of
the eight future frames, $\vect{x}_{0}=\Set{\vect{x}^{j}_{t=0}}_{j=i+1}^{i+8}\in\mathbb{R}^{8\times4\times8\times8}$,
and the eight future clear-sky-index values $\vect{y}_0=[y_{t=0}^j]\in\mathbb{R}^{8}$, a horizon of
sixteen minutes (superscript $j$: time step; subscript $t$: diffusion step):
\begin{equation}
\big(\vect{x}_{0},\,\vect{y}_{0}\big) = F_\Theta(\vect{a},\,\vect{e}).
\label{eq:forecastmap}
\end{equation}
The clear-sky index (CSI) is $k=\mathrm{GHI}/\max(I_{clr},1)$; the diffusion
target is $k$ clipped to $[0,1.67]$ and standardized, $y=(k-\mu_k)/\sigma_k$ with frozen constants $\mu_k=0.7256$ and $\sigma_k=0.3822$, and the
forecast GHI is the inverted $k$ multiplied by the future $I_{clr}$. \textbf{The model receives no
measured irradiance history}: $I_{clr}$ is deterministic given time and location, and nothing else
about the past irradiance series reaches the model (Section~\ref{sec:inputs} ablates exactly this).

Frames are encoded by a Stable Diffusion VAE, one $4\times8\times8$ latent per $3\times64\times64$
frame, and the image branch diffuses in that latent space, $\vect{x}_t\in\mathbb{R}^{8\times4\times8\times8}$;
the CSI sequence is diffused directly, $\vect{y}_t\in\mathbb{R}^{8}$. Both share one diffusion timestep
$t$ of $T=1000$ training steps, so they carry the same noise level, and Eq.~\eqref{eq:x0hat} gives a
denoised estimate of each: $\hat{\vect{x}}_0$ of the future sky and $\hat{\vect{y}}_0$ of the future
CSI. The proposed model's irradiance head reads $\hat{\vect{x}}_0$ (Section~\ref{sec:method-step});
$\hat{\vect{y}}_0$ is read only by the cross-modal variant (Appendix~\ref{sec:app-designspace}).

\subsection{Framework}
\label{sec:method-arch}

\begin{figure}[pos=!htbp]
\centering
\includegraphics[width=\textwidth, trim={1cm, 0, 1cm, 0}]{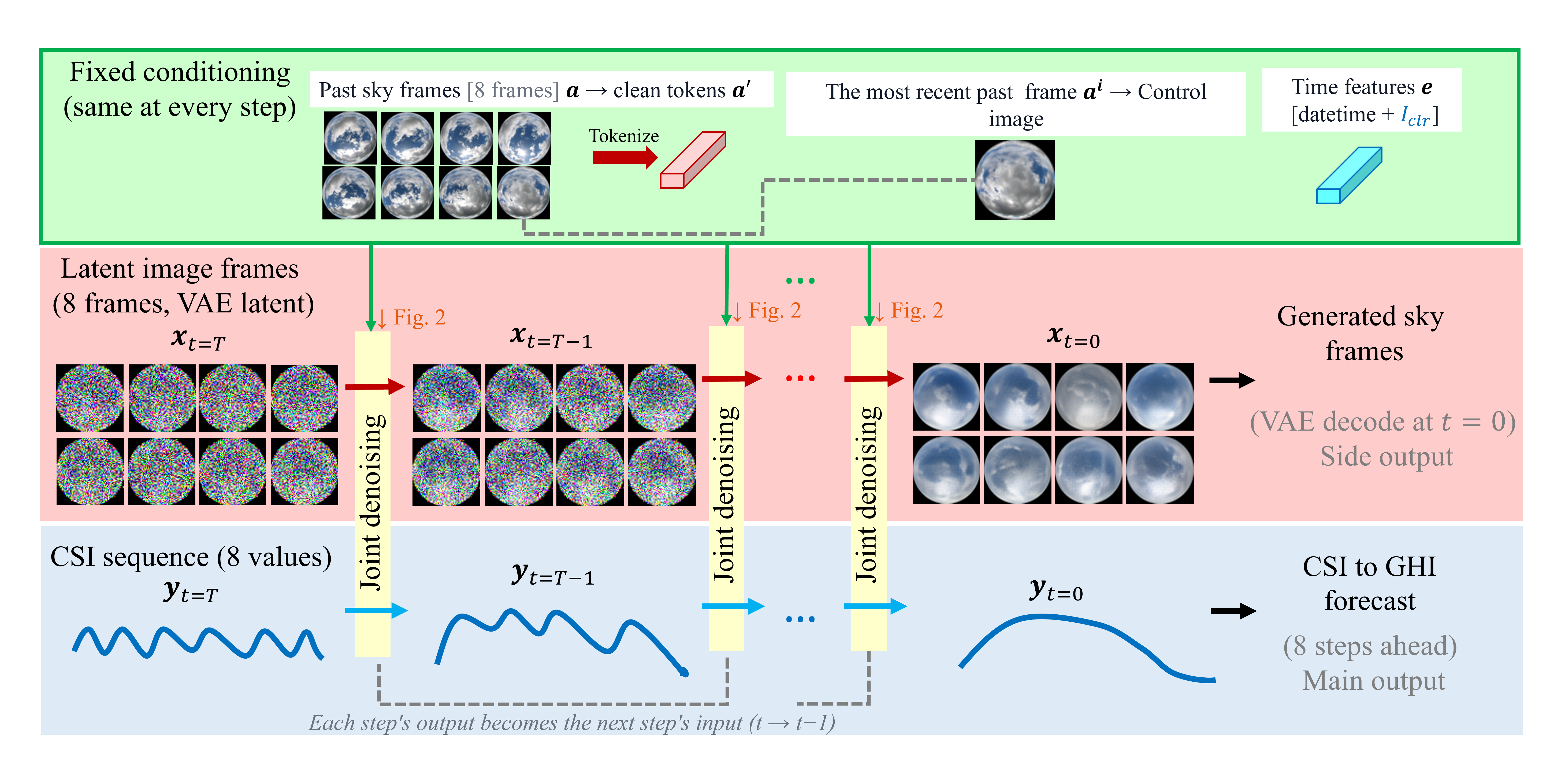}
\caption{Joint reverse diffusion: the 50-step sampling trajectory. Both streams are denoised
  together at every step. The fixed conditioning (green) is identical at every step. The image
  latents (pink) run from noise to the decoded future frames; the clear-sky-index sequence (blue) is
  denoised in parallel in standardized value space and converted to an irradiance forecast. The sky
  frames are a side output --- the forecast is carried by the clear-sky-index stream. The frames shown are
  the ramp-up case of Figure~\ref{fig:frames}: its eight input frames and one sample of the proposed model's
  output. The noisy frames, made by adding noise to that output, and the clear-sky-index curves are
  illustrative. One step is expanded in
  Figure~\ref{Fig:diagram2}.}\label{Fig:diagram1}
\end{figure}

$F_\Theta(\cdot)$, with $198{,}834{,}269$ parameters (Figure~\ref{Fig:diagram1}), comprises
\begin{itemize}
	\item the \emph{Stable Diffusion VAE encoder} $\mathcal{E}$, applied frame by frame, fine-tuned on sky
	images in the first training stage and frozen thereafter (its decoder $\mathcal{D}$ maps latents back
	to frames);
	\item an \emph{image tokenizer} $\mathcal{T}:\mathbb{R}^{4\times8\times8}\to\mathbb{R}^{16\times256}$,
	so that the eight frames of a clip become $128$ tokens;
	\item an \emph{irradiance head} $H_\phi$, a CSI denoiser: a transformer over the
	eight CSI positions that self-attends across them and cross-attends to a sequence of image tokens;
	\item an \emph{image branch} $G_\theta$, an image denoiser: the Stable Diffusion UNet, a ControlNet branch, and a
	temporal refiner that corrects the per-frame noise estimates jointly across the eight future frames.
\end{itemize}
Appendix~\ref{sec:app-arch} gives their dimensions. Three conditioning inputs are computed once from
the past and held fixed: the \emph{clean tokens} of the past frames,
$\vect{a}' = \mathcal{T}(\mathcal{E}(\vect{a}))\in\mathbb{R}^{128\times256}$; the \emph{control
image}, the most recent frame $\vect{a}^i$ in pixel space, passed to the ControlNet; and the
\emph{time features} $\vect{e}$. The future latents $\vect{x}_t$ and CSI sequence $\vect{y}_t$ start as
noise at $t=T$ and are jointly denoised by $G_\theta$ and $H_\phi$ over $50$ reverse steps.

\begin{figure*}[pos=!htbp]
\centering
\includegraphics[width=\textwidth]{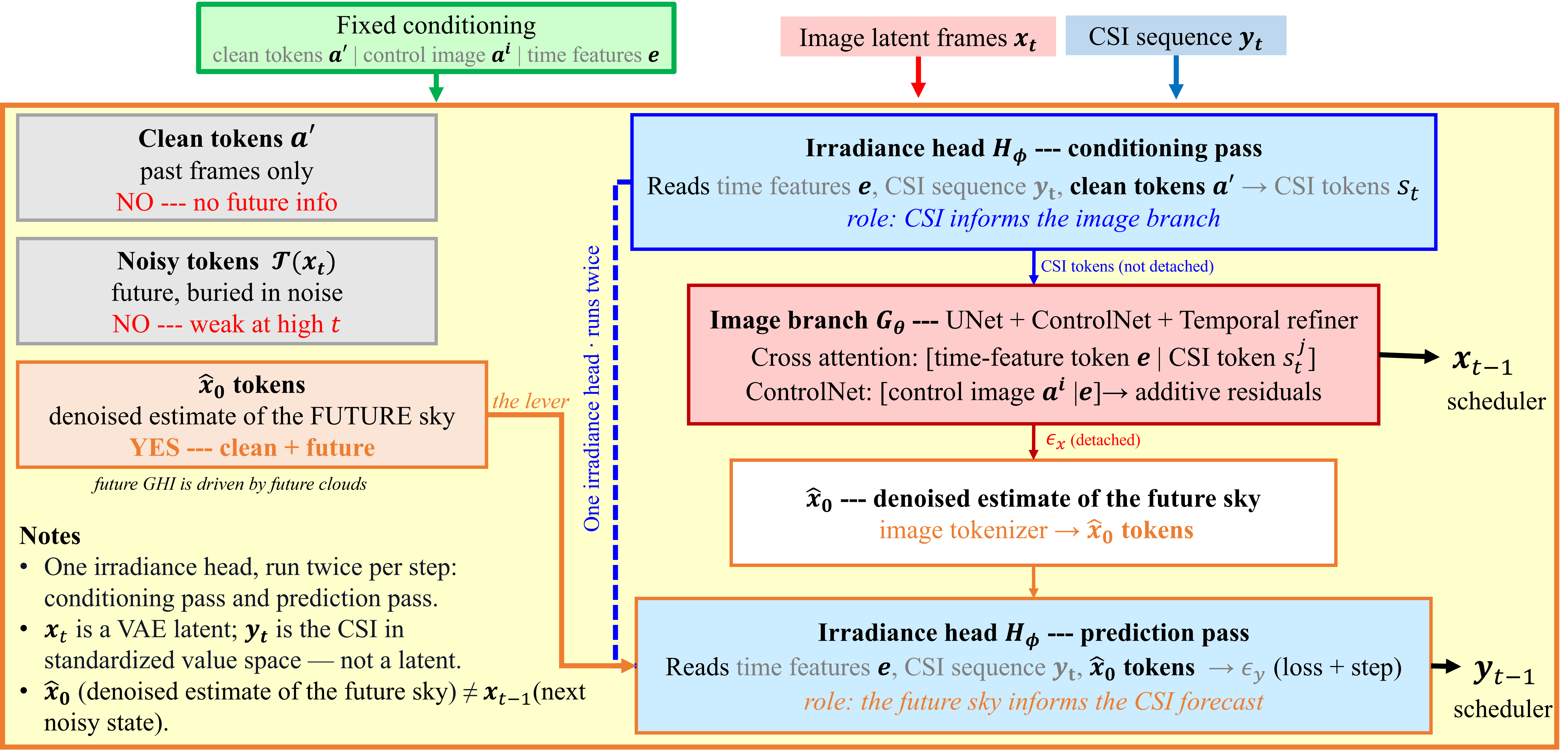}
\caption{Inside one denoising step of the proposed model, and why the irradiance head reads
  $\hat{\vect{x}}_0$. A single irradiance head is run twice. The conditioning pass emits CSI tokens, one
  per future frame, that enter the image branch's cross-attention context \emph{undetached},
  alongside the time-feature token; the control image enters separately through ControlNet as
  additive residuals, and is not a cross-attention token. The image branch's $\epsilon_x$ then
  yields $\hat{\vect{x}}_0$, a denoised estimate of the \emph{future} sky, which the prediction pass
  reads. Because $\hat{\vect{x}}_0$ is built from a detached $\epsilon_x$, the image-to-CSI link
  carries forward values only: no gradient cycle closes. The four boxes on the right implement
  Eqs.~\eqref{eq:condpass}, \eqref{eq:imgbranch}, \eqref{eq:x0hatimg} and \eqref{eq:predpass} in
  order; the conditioning pass's provisional noise estimate $\epsilon'_y$, unused by this model, is
  not drawn. The candidate inputs on the left are three of the four choices in
  Eq.~\eqref{eq:axis}.}\label{Fig:diagram2}
\end{figure*}

\subsection{One denoising step}
\label{sec:method-step}

$H_\phi$ and $G_\theta$ share the timestep $t$ and the fixed conditioning, and run in sequence,
$H_\phi \rightarrow G_\theta \rightarrow H_\phi$. A \emph{conditioning pass} informs the image branch
of the current CSI state, and a \emph{prediction pass} of the same irradiance head reads the image
branch's denoised estimate of the future sky (Figure~\ref{Fig:diagram2}).

\paragraph{Conditioning pass.} $H_\phi$ reads the noisy CSI sequence $\vect{y}_t$, the time features
$\vect{e}$ and the clean tokens $\vect{a}'$, and returns a provisional noise estimate
$\epsilon'_y\in\mathbb{R}^{8}$ and \emph{CSI tokens} $s_t=\Set{s^j_t}$, one per future frame,
$s^j_t\in\mathbb{R}^{4}$:
\begin{equation}
\big(\epsilon'_y,\, s_t\big) = H_\phi\big(\vect{y}_t,\,t;\ \vect{e},\ \vect{a}'\big).
\label{eq:condpass}
\end{equation}
The proposed model uses only $s_t$; $\epsilon'_y$ is used by the cross-modal variant.

\paragraph{Imagery pass.} $G_\theta$ predicts the noise $\epsilon_x$ in the future-frame latents, of
the shape of $\vect{x}_t$, in two steps. The UNet and the ControlNet first estimate the noise
$\tilde{\epsilon}^j$ of each future frame separately, and the temporal refiner $R$ then corrects the
eight estimates jointly:
\begin{equation}
\begin{gathered}
\tilde{\epsilon}^j = \mathrm{UNet}\big(\vect{x}^j_t,\,t;\ [\,\vect{e},\,s^j_t\,],\
\mathrm{ControlNet}(\vect{x}^j_t,\,t;\ \vect{a}^i,\,\vect{e})\big),\\
\epsilon_x^j = \tilde{\epsilon}^j + R^j\big(\vect{x}_t,\,\tilde{\epsilon}\big),
\qquad j=i+1,\dots,i+8,
\end{gathered}
\label{eq:imgbranch}
\end{equation}
where $R^j$ is the temporal refiner's correction for frame $j$, computed from all eight frames, and we write the
whole pass as $\epsilon_x = G_\theta\big(\vect{x}_t,\,t;\ [\,\vect{e},\,s_t\,],\ \vect{a}^i\big)$. The
UNet cross-attends to $[\,\vect{e},\,s^j_t\,]$, the ControlNet reads the control image $\vect{a}^i$, and
$R$ attends across the eight frames only (Appendix~\ref{sec:app-arch}). Given $\epsilon_x$,
the model forms the \emph{denoised estimate} of the future sky by Eq.~\eqref{eq:x0hat}; for
$j = i+1,\dots,i+8$,
\begin{equation}
\hat{\vect{x}}_0^j = \frac{\vect{x}^j_t - \sqrt{1-\bar\alpha_t}\;\mathrm{sg}[\epsilon_x^j]}{\sqrt{\bar\alpha_t}},
\label{eq:x0hatimg}
\end{equation}
where $\mathrm{sg}[\cdot]$ is the stop-gradient operator.

\paragraph{Prediction pass.} The irradiance head runs again, now reading the tokens of the denoised
estimate, $\mathcal{T}(\hat{\vect{x}}_0)\in\mathbb{R}^{128\times256}$, in place of the clean tokens, and
outputs the noise estimate $\epsilon_y\in\mathbb{R}^{8}$ that bears the irradiance loss:
\begin{equation}
\epsilon_y = H_\phi\big(\vect{y}_t,\,t;\ \vect{e},\ \mathcal{T}(\hat{\vect{x}}_0)\big).
\label{eq:predpass}
\end{equation}
\paragraph{Learning loss.} $\epsilon_y$ drives the CSI scheduler step and $\epsilon_x$ the image
scheduler step. Training minimizes
\begin{equation}
\mathcal{L} = \mathbb{E}\Big[\,\big\|\epsilon_x^{*}-\epsilon_x\big\|_2^2 + \lambda\,\big\|\epsilon_y^{*}-\epsilon_y\big\|_2^2\,\Big],
\qquad \lambda = 1,
\label{eq:loss}
\end{equation}
where $\epsilon_x^{*}$ and $\epsilon_y^{*}$ are the noises added to each stream in
Eq.~\eqref{eq:forward}.

\paragraph{Gradient topology.} The CSI tokens enter the image branch \emph{undetached}, so the irradiance
head receives gradient from the image loss. In the other direction $\hat{\vect{x}}_0$ is built
from a \emph{detached} $\epsilon_x$ (Eq.~\eqref{eq:x0hatimg}), so the image-to-CSI link carries forward
values only: the irradiance loss reaches the image tokenizer but not the UNet, the ControlNet or the
temporal refiner. No gradient cycle closes, and the image branch is optimized for frame reconstruction
alone, the point on which this design differs from the closest concurrent method~\citep{huang2026pvmm}.

\paragraph{Design choice.} The prediction pass, Eq.~\eqref{eq:predpass}, reads the tokens of the
denoised estimate. The design choice ablated in Section~\ref{sec:ablation-source} is this input:
replacing $\mathcal{T}(\hat{\vect{x}}_0)$ in Eq.~\eqref{eq:predpass} by a general input $\vect{v}$ gives
\begin{equation}
\epsilon_y = H_\phi\big(\vect{y}_t,\,t;\ \vect{e},\ \vect{v}\big),\qquad
\vect{v} \in \Big\{\,
\underbrace{\vect{a}'}_{\text{clean}},\
\underbrace{\mathcal{T}(\vect{x}_t)}_{\text{noisy}},\
\underbrace{[\,\mathcal{T}(\vect{x}_t),\,\vect{a}'\,]}_{\text{concat}},\
\underbrace{\mathcal{T}(\hat{\vect{x}}_0)}_{\text{proposed}}
\,\Big\}.
\label{eq:axis}
\end{equation}
The clean tokens carry no information about the future sky; the noisy tokens carry it, but weakly at
high $t$ and differently at training and sampling time; the denoised estimate is both clean and about
the future. \emph{Concat} ($256$ tokens against $128$) has the information set of \PVMM{}, which
conditions on the noised iterates of both modalities and forms no denoised estimate. Only the proposed
input needs two passes of the irradiance head, and Section~\ref{sec:ablation-source} measures all four.

\paragraph{Cross-modal variant.} We also evaluate the reverse path, in which a denoised estimate of the
future CSI is fed back into the image branch (Appendix~\ref{sec:app-designspace}). Each variant, like
each input in Eq.~\eqref{eq:axis}, is a separately trained model.

\subsection{Training and sampling}
\label{sec:training}
\label{sec:method-training}

Training runs five stages in sequence: three train the VAE, the UNet and the ControlNet one at a time,
and two train jointly, with both streams noised at a shared timestep. Phase~1 of joint training trains only the modules that start from random initialization. Phase~2
fine-tunes all modules except the VAE at a hundredth of the phase-1 learning rate, so that the
pretrained image branch adapts rather than being overwritten
(Table~\ref{tab:training}; checkpoint and budget conventions in Appendix~\ref{sec:app-protocol}).

At inference both streams are co-denoised over $50$ reverse steps with the conditioning fixed; at
$t=0$ the latents are decoded to frames and the CSI sequence is mapped back to irradiance. The frames
are a by-product: the forecast is read from the irradiance head, with no frame-to-irradiance regression
stage anywhere in the pipeline.

%% file: sec_setup.tex
\section{Data and experimental setup}
\label{sec:setup}

\subsection{Dataset}
\label{sec:data}

We use sky images and global horizontal irradiance (GHI) from the SIRTA atmospheric
observatory~\citep{haeffelin2005sirta} in Palaiseau, France (48.713$^\circ$N, 2.208$^\circ$E), a site
whose imagery and irradiance records are widely used for sky-image
forecasting~\citep{varaschin2025benchmarking}. Daytime sky images at two-minute cadence
are resized to $64\times64$ pixels and aligned with the GHI measurement and the clear-sky irradiance
$I_{clr}$ at the same timestamp, computed with the Ineichen--Perez clear-sky
model~\citep{ineichen2002linke} as implemented in pvlib~\citep{holmgren2018pvlib}. A sample is a sequence of eight past frames and the eight following
frames, $j=i-7,\dots,i+8$, so that a forecast covers a sixteen-minute horizon
(Section~\ref{sec:method-setup}).

\paragraph{In-period data (2023).} All models in the main comparison are trained and evaluated on
calendar year 2023. Samples are partitioned $0.80/0.10/0.10$ into training, validation and test sets;
the test set contains $12{,}906$ sequences from $43$ days. Within each month, the earliest tenth of
the samples in time order forms the test set and the next tenth the validation set, so the split falls
almost entirely between days. We quantify the remaining adjacency. Fewer than $2\%$ of test sequences have a training sequence from
the same day within $\pm1$~h, and fewer than $2\%$ a validation sequence; $88\%$ of the matched subset
(below) falls on the $34$ of $43$ test days that contain neither.

\paragraph{Cross-year data (2021--2022 $\rightarrow$ 2023).} To remove that adjacency altogether, a
second set of models is trained and validated on 2021--2022 only, with a day-blocked split, and tested
on the same 2023 test sequences. No training or validation sample shares a calendar year, let alone a
day, with a test sample. Any difference from the in-period result therefore reflects mainly the shift in
training distribution (Section~\ref{sec:crossyear}).

\paragraph{Matched test subset.} Cross-method comparisons use a subset of $1998$ test sequences,
stratified across sky conditions and drawn from $40$ of the $43$ test days. Every method is scored on exactly these inputs, and the smart-persistence reference
is recomputed on the same rows. Of the $1998$ scored test windows, $13$ are not contiguous. In five, the input and target frames fall
on different days, separated by a night. In the other eight, all frames are from the same day but one
or two frames are missing, so the spacing is four or six minutes at some point in the window. Training windows are
affected at a similar rate. We keep them, since every method is scored on the same windows; excluding them
changes no ranking (Table~\ref{tab:contiguity}). SIRTA has since reviewed the 2023 irradiance record and
marked as unusable some minutes that were provisional in our copy; $62$ of the $1998$ samples have such a
minute among their targets or their persistence origin, and excluding them also changes no ranking.

\subsection{Experimental protocol}
\label{sec:protocol}

Each experiment in Sections~\ref{sec:comparison}--\ref{sec:ablations} and Appendix~\ref{sec:app-designspace} uses the data as follows.

\begin{figure}[pos=!htbp]
	\centering
	\includegraphics[width=0.5\textwidth]{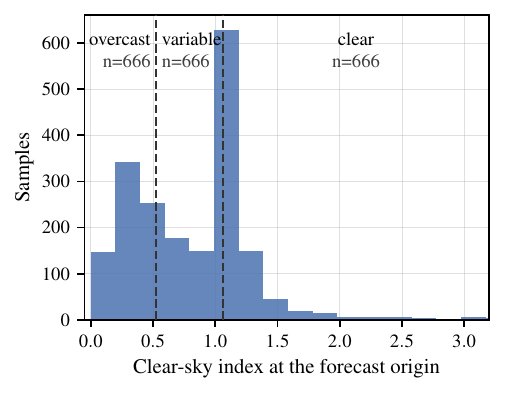}
	\caption{Sky-condition tertiles on the matched test subset: the distribution of the clear-sky index
		$k$ at the forecast origin. Dashed lines are the tertile edges, $k=0.527$ and $k=1.063$, and each
		region holds $666$ samples. The horizontal axis is cut at $k=3.2$; the $43$ samples above it are
		binned but not shown. Every sample with $k>2$ has a solar elevation below $10^\circ$ at the
		origin, where the clear-sky model is least reliable.}
	\label{fig:regimes}
\end{figure}

\begin{itemize}
\item \textbf{Comparison (Section~\ref{sec:comparison}, Table~\ref{tab:comparison}).} Every method is
trained on the 2023 training set and scored on the matched subset. Confidence intervals are $95\%$
day-block bootstrap intervals: the $40$ test days are resampled with replacement ($5000$ draws), every
score is recomputed on each draw, and paired contrasts between two methods use the same draws.
\item \textbf{Cross-year retraining (Section~\ref{sec:crossyear}, Table~\ref{tab:crossyear}).} Six
methods, the proposed model among them, are retrained on the 2021--2022 data and scored on the same
matched subset and reference as the comparison. This is the paper's leak-free test of the main result.

\item \textbf{Frozen regressor (Sections~\ref{sec:comparison}--\ref{sec:imagequality}).} A U-Net regressor ($306{,}120$ parameters) that maps a single sky image to its concurrent GHI is
trained on \emph{real} sky images and then frozen. Its training data are the $107{,}454$ single frames
of the 2023 training split, downsampled from $256\times256$ to $64\times64$; it uses a
mean-absolute-error loss on GHI and is selected by early stopping on the validation split. On the frames of the test split it reaches an RMSE of $57.1$~W\,m$^{-2}$. It
reads one $64\times64$ frame at a time, with no covariates, and returns GHI directly, without
clipping; it never sees a generated frame during training. Section~\ref{sec:comparison} uses it to
score the methods that produce frames but no irradiance forecast of their own;
Section~\ref{sec:cascade} applies it to the true, reference and generated frames, and
Section~\ref{sec:imagequality} to the frames of every frame-producing method.

\item \textbf{Per-horizon, sky-condition and ramp-event performance (Section~\ref{sec:calibration}).} On the matched subset,
sky-condition regimes are the tertiles of the clear-sky index $k$ at the forecast origin (edges
$k=0.527$ and $k=1.063$; $666$ samples each; Figure~\ref{fig:regimes}), referred to as overcast, variable, and clear, from the lowest to the highest tertile;
the labels describe the level of $k$ at the origin, not measured variability, since an intermediate
$k$ usually arises from broken cloud but also from a steady thin overcast. A sample is a ramp if GHI
changes by at least $200$~W\,m$^{-2}$ between two consecutive forecast steps ($n=334$); a tertile
therefore describes the state at the origin and a ramp the change that follows. The highest tertile is not a clear-sky
class: it also collects windows near sunrise and sunset, where the clear-sky model is least reliable.
Half of the lowest and $47\%$ of the highest tertile have a solar elevation below $20^\circ$ at the
origin, and every one of the $68$ windows with $k>2$ lies below $10^\circ$. Excluding the windows
below $10^\circ$ or $20^\circ$ leaves the ranking of the methods of Table~\ref{tab:comparison}
unchanged, except that the two DLinear variants, $0.002$ apart in CRPSS at that cut, swap places at $20^\circ$;
it changes the sign of no tested paired comparison (Appendix~\ref{sec:app-protocol},
Table~\ref{tab:elevation}).
\item \textbf{Ablations (Section~\ref{sec:ablations}).} Arms are trained on the 2023 training set
with the stages of Section~\ref{sec:method-training} (budget exceptions in Appendix~\ref{sec:app-protocol}) and scored on the matched subset; the
irradiance-stream ablation pairs each of two seeds with its own control.

\item \textbf{Cross-modal variants (Appendix~\ref{sec:app-designspace}).} Three arms that feed the
denoised estimate of the future CSI back into the image branch are scored like the ablations, on the matched
subset.
\end{itemize}

All joint arms in the comparisons of Sections~\ref{sec:comparison}--\ref{sec:ablations} sample $50$
reverse diffusion steps and form an $M=10$-member ensemble, each member from an independent initial
noise; both settings are the same for every one of those arms and every seed. No post-hoc
calibration of the predictive distributions is applied anywhere in this paper; all intervals reported here
are raw model output, since a calibration layer would improve interval scores without informing the
architectural comparison.

\paragraph{Run-to-run variation.}
\label{sec:noise}
Three sources of run-to-run variation were measured before any comparison was read, and differences
below them are not interpreted. (i)~\emph{Training:} the proposed model's three training seeds span
$1.08$~W\,m$^{-2}$ in pooled RMSE. (ii)~\emph{Sampling:} re-running the sampler with a different state
of its noise generator moves RMSE by $1.34$~W\,m$^{-2}$, forecast skill by $0.011$ and CRPSS by
$0.006$. (iii)~\emph{Seeding convention:} drawing the sampler noise once per position within a test
batch (Table~\ref{tab:comparison}) rather than once per test sample (the paired ablations of
Section~\ref{sec:ablation-source}) moves pooled RMSE by $0.5$--$1.4$~W\,m$^{-2}$, depending on the
training seed. The day-block bootstrap intervals reflect only which test days are scored and include
none of these sources. One score is much noisier: the frozen regressor's RMSE on an arm's generated
frames varies by $3.48$~W\,m$^{-2}$ across seeds of one arm but by up to $20.3$~W\,m$^{-2}$ between
separately trained runs, so for that score we interpret only the sign of a paired difference. The
other differences interpreted in this paper exceed the three scales above, most of them several-fold.
The two smallest, the point-accuracy gap to \SunSet{} ($6.5$~W\,m$^{-2}$) and the cost of the
cross-modal path ($4.5$~W\,m$^{-2}$, Section~\ref{sec:ablation-source}), are noted where they arise.

\subsection{Evaluation metrics}
\label{sec:metrics}

Let $y_{n,h}$ be the observed GHI of test sample $n=1,\dots,N$ at forecast step $h=1,\dots,8$, and
$\hat y^{(m)}_{n,h}$, $m=1,\dots,M$, the members of a probabilistic forecast, with point forecast
$\hat y_{n,h}$ their mean. Every score is pooled over all $NH$ sample--step pairs before any ratio is
taken; skill scores are a ratio of two pooled quantities, not a mean of per-step ratios.

\begin{itemize}
	\item \textbf{Point accuracy.} The root-mean-square error and the forecast skill against smart persistence
	(SPM) are
	\begin{equation}
		\mathrm{RMSE} = \sqrt{\frac{1}{NH}\sum_{n,h}\big(\hat y_{n,h}-y_{n,h}\big)^2},\qquad
		\mathrm{FS} = 1-\frac{\mathrm{RMSE}}{\mathrm{RMSE}_{\mathrm{SPM}}}.
		\label{eq:rmse}
	\end{equation}
	SPM holds the clear-sky index at the forecast origin $i$ constant over the horizon,
	\begin{equation}
		\hat y^{\mathrm{SPM}}_{n,h}=k_{n}\,I_{clr,n,h},\qquad k_n=\frac{\mathrm{GHI}_{n,i}}{\max(I_{clr,n,i},1)},
		\label{eq:spm}
	\end{equation}
	and is recomputed on the same samples as the method being scored.
	
	\item \textbf{Probabilistic accuracy.}  The continuous ranked probability score uses the fair estimator for
	a finite ensemble,
	\begin{equation}
		\mathrm{CRPS} = \frac{1}{NH}\sum_{n,h}\Bigg[\frac{1}{M}\sum_{m}\big|\hat y^{(m)}_{n,h}-y_{n,h}\big|
		-\frac{1}{2M(M-1)}\sum_{m,m'}\big|\hat y^{(m)}_{n,h}-\hat y^{(m')}_{n,h}\big|\Bigg],\qquad
		\mathrm{CRPSS} = 1-\frac{\mathrm{CRPS}}{\mathrm{MAE}_{\mathrm{SPM}}}.
		\label{eq:crps}
	\end{equation}
	where $m$ and $m'$ both index the $M$ ensemble members; the $m=m'$ terms vanish, and normalizing the
	remaining $M(M-1)$ ordered pairs is what makes the estimator fair. For a deterministic forecast the CRPS reduces to the mean absolute error (MAE), which is why the
	reference in the CRPS skill score (CRPSS) is the MAE of SPM.
	
	\item \textbf{Reliability.} The prediction-interval coverage probability at nominal level $\alpha$ is the
	fraction of observations inside the central interval spanned by the empirical
	$\tfrac{1-\alpha}{2}$ and $\tfrac{1+\alpha}{2}$ quantiles, $\ell_{n,h}$ and $u_{n,h}$, of the $M$ members. The spread-skill ratio (SSR) compares the error of the ensemble mean with the
ensemble's own spread,
	\begin{equation}
		\mathrm{PICP}_\alpha = \frac{1}{NH}\sum_{n,h}\mathbf{1}\big[\ell_{n,h}\le y_{n,h}\le u_{n,h}\big],
		\qquad
		\mathrm{SSR} = \frac{\mathrm{RMSE}}{\sqrt{\frac{1}{NH}\sum_{n,h}\sigma^2_{n,h}}},
		\label{eq:picp}
	\end{equation}
	where $\sigma^2_{n,h}=\frac{1}{M-1}\sum_{m}\big(\hat y^{(m)}_{n,h}-\hat y_{n,h}\big)^2$ is the unbiased variance of the members about their mean. For $M$ exchangeable draws from a
	calibrated forecast, the expected SSR is $\sqrt{(M+1)/M}\approx1.05$ at $M=10$, not $1$. Values above
	the target indicate an ensemble whose spread is too small for its errors and below it one whose spread
	is too large. Following the tolerance used in scoring, $\pm10\%$ of the target, an ensemble is called
	under-dispersed above $1.1\sqrt{(M+1)/M}$ and over-dispersed below $0.9\sqrt{(M+1)/M}$
	($1.154$ and $0.944$ at $M=10$), and roughly calibrated in between. PICP and SSR are undefined for deterministic forecasts.
	
\end{itemize}

%% file: sec_comparison.tex
\section{Performance comparison}
\label{sec:comparison}

Table~\ref{tab:comparison} compares the proposed model with four groups of baselines, listed below
in the order in which they appear in the table. All methods are scored on the matched test subset
against the same smart-persistence reference, Eq.~\eqref{eq:spm} (pooled
$\mathrm{RMSE}=127.37$~W\,m$^{-2}$, $\mathrm{MAE}=56.08$~W\,m$^{-2}$). (i)~\emph{Scalar forecasters}:
DLinear~\citep{zeng2023dlinear}, PatchTST~\citep{nie2023patchtst} and TimeXer~\citep{wang2024timexer},
as implemented in the Time-Series-Library~\citep{tslib}, which forecast from the measured irradiance history
alone, first without and then with the clear-sky irradiance $I_{clr}$ as an additional input.
(ii)~A \emph{deterministic sky-image network}, \SunSet{}~\citep{sun2019sunset}, which reads the past
sky images together with the irradiance history and returns a point forecast; the per-horizon
figures also show it without the clear-sky input, a variant not tabulated here. (iii)~\emph{Video
predictors}, \ConvLSTM{}~\citep{shi2015convlstm}, \PhyDNet{}~\citep{leguen2020phydnet} and
\PhyDNetGAN{}~\citep{leguen2020phydnet,nie2024skygpt},
which forecast future frames only. Their irradiance is read from the generated frames by the frozen
regressor of Section~\ref{sec:protocol}, the frame at lead time $j$ giving the forecast for $i{+}j$;
being point forecasts, their CRPS equals their MAE. (iv)~\emph{Published generative forecasters}: \SkyGPT{}~\citep{nie2024skygpt}, a stochastic video
predictor, and the model of \citet{xiong2025multimodal}. Each of the ten sampled futures of \SkyGPT{} is
passed through the same regressor, and its forecast is the mean of the ten, as for every ensemble in
this paper. The model of \citet{xiong2025multimodal} couples a stochastic video model to its own
probabilistic forecasting head, trained separately from the video model, that also reads the
irradiance history and temporal covariates. Its forecasts are taken from that head, not from the
regressor. (v)~This work:
\propFuture{} and \propPast{}. Implementation details and the fidelity checks applied to each
reimplementation are given in Appendix~\ref{sec:app-baselines}.

The comparison is not symmetric in its inputs. The proposed model receives no measured irradiance: it
conditions on past sky images and on deterministic covariates at the forecast origin, whereas the
scalar forecasters, \SunSet{} and the model of \citet{xiong2025multimodal} also read the irradiance
history; Appendix~\ref{sec:lattice} lists the information each method conditions on.
Section~\ref{sec:inputs} quantifies the effect of this difference.
The confidence intervals are day-block bootstrap intervals (Section~\ref{sec:protocol}) and therefore
reflect variation across test days, not across training runs.

\begin{table*}[t]
\centering
\caption{Forecast accuracy on the matched test subset ($N=1998$), with $95\%$ day-block bootstrap
intervals. RMSE and CRPS in W\,m$^{-2}$; FS and CRPSS are skill scores against smart persistence,
higher is better. Rules separate, from top to bottom: smart persistence; scalar forecasters without
and with $I_{clr}$; the deterministic sky-image network; video predictors scored through the frozen
regressor; published generative forecasters; and this work. \underline{\textbf{Bold underlined}}: best in each column; \textbf{bold}: second best (smart persistence, the reference, is not ranked).}
\label{tab:comparison}
\footnotesize
\setlength{\tabcolsep}{2pt}
\begin{tabular}{@{}Q{\wMethod}P{\wRMSE}P{\wFS}P{\wCRPS}P{\wCRPSS}@{}}
\toprule
Method & RMSE\,\arrowdown{} [95\% CI] & FS\,\arrowup{} [95\% CI] & CRPS\,\arrowdown{} [95\% CI] & CRPSS\,\arrowup{} [95\% CI] \\
\midrule
\DTLforeach*[\equal{\Grp}{0}]{comp}{\Grp=group,\Mth=method,\RM=rmse,\RMc=rmseci,\FS=fs,\FSc=fsci,\CR=crps,\CRc=crpsci,\CS=crpss,\CSc=crpssci,\Brm=brmse,\Bfs=bfs,\Bcr=bcrps,\Bcs=bcrpss}{%
  \DTLiffirstrow{}{\\}\csname\Mth\endcsname & \csname\Brm\endcsname{\RM}~\RMc & \csname\Bfs\endcsname{\FS}~\FSc & \csname\Bcr\endcsname{\CR}~\CRc & \csname\Bcs\endcsname{\CS}~\CSc}\\
\midrule
\DTLforeach*[\equal{\Grp}{1}]{comp}{\Grp=group,\Mth=method,\RM=rmse,\RMc=rmseci,\FS=fs,\FSc=fsci,\CR=crps,\CRc=crpsci,\CS=crpss,\CSc=crpssci,\Brm=brmse,\Bfs=bfs,\Bcr=bcrps,\Bcs=bcrpss}{%
  \DTLiffirstrow{}{\\}\csname\Mth\endcsname & \csname\Brm\endcsname{\RM}~\RMc & \csname\Bfs\endcsname{\FS}~\FSc & \csname\Bcr\endcsname{\CR}~\CRc & \csname\Bcs\endcsname{\CS}~\CSc}\\
\midrule
\DTLforeach*[\equal{\Grp}{2}]{comp}{\Grp=group,\Mth=method,\RM=rmse,\RMc=rmseci,\FS=fs,\FSc=fsci,\CR=crps,\CRc=crpsci,\CS=crpss,\CSc=crpssci,\Brm=brmse,\Bfs=bfs,\Bcr=bcrps,\Bcs=bcrpss}{%
  \DTLiffirstrow{}{\\}\csname\Mth\endcsname & \csname\Brm\endcsname{\RM}~\RMc & \csname\Bfs\endcsname{\FS}~\FSc & \csname\Bcr\endcsname{\CR}~\CRc & \csname\Bcs\endcsname{\CS}~\CSc}\\
\midrule
\DTLforeach*[\equal{\Grp}{3}]{comp}{\Grp=group,\Mth=method,\RM=rmse,\RMc=rmseci,\FS=fs,\FSc=fsci,\CR=crps,\CRc=crpsci,\CS=crpss,\CSc=crpssci,\Brm=brmse,\Bfs=bfs,\Bcr=bcrps,\Bcs=bcrpss}{%
  \DTLiffirstrow{}{\\}\csname\Mth\endcsname & \csname\Brm\endcsname{\RM}~\RMc & \csname\Bfs\endcsname{\FS}~\FSc & \csname\Bcr\endcsname{\CR}~\CRc & \csname\Bcs\endcsname{\CS}~\CSc}\\
\midrule
\DTLforeach*[\equal{\Grp}{4}]{comp}{\Grp=group,\Mth=method,\RM=rmse,\RMc=rmseci,\FS=fs,\FSc=fsci,\CR=crps,\CRc=crpsci,\CS=crpss,\CSc=crpssci,\Brm=brmse,\Bfs=bfs,\Bcr=bcrps,\Bcs=bcrpss}{%
  \DTLiffirstrow{}{\\}\csname\Mth\endcsname & \csname\Brm\endcsname{\RM}~\RMc & \csname\Bfs\endcsname{\FS}~\FSc & \csname\Bcr\endcsname{\CR}~\CRc & \csname\Bcs\endcsname{\CS}~\CSc}\\
\midrule
\DTLforeach*[\equal{\Grp}{5}]{comp}{\Grp=group,\Mth=method,\RM=rmse,\RMc=rmseci,\FS=fs,\FSc=fsci,\CR=crps,\CRc=crpsci,\CS=crpss,\CSc=crpssci,\Brm=brmse,\Bfs=bfs,\Bcr=bcrps,\Bcs=bcrpss}{%
  \DTLiffirstrow{}{\\}\csname\Mth\endcsname & \csname\Brm\endcsname{\RM}~\RMc & \csname\Bfs\endcsname{\FS}~\FSc & \csname\Bcr\endcsname{\CR}~\CRc & \csname\Bcs\endcsname{\CS}~\CSc}\\
\midrule
\DTLforeach*[\equal{\Grp}{6}]{comp}{\Grp=group,\Mth=method,\RM=rmse,\RMc=rmseci,\FS=fs,\FSc=fsci,\CR=crps,\CRc=crpsci,\CS=crpss,\CSc=crpssci,\Brm=brmse,\Bfs=bfs,\Bcr=bcrps,\Bcs=bcrpss}{%
  \DTLiffirstrow{}{\\}\csname\Mth\endcsname & \csname\Brm\endcsname{\RM}~\RMc & \csname\Bfs\endcsname{\FS}~\FSc & \csname\Bcr\endcsname{\CR}~\CRc & \csname\Bcs\endcsname{\CS}~\CSc}\\
\bottomrule
\end{tabular}
\end{table*}

\begin{figure}[pos=!htbp]
  \centering
  \includegraphics[width=0.8\linewidth]{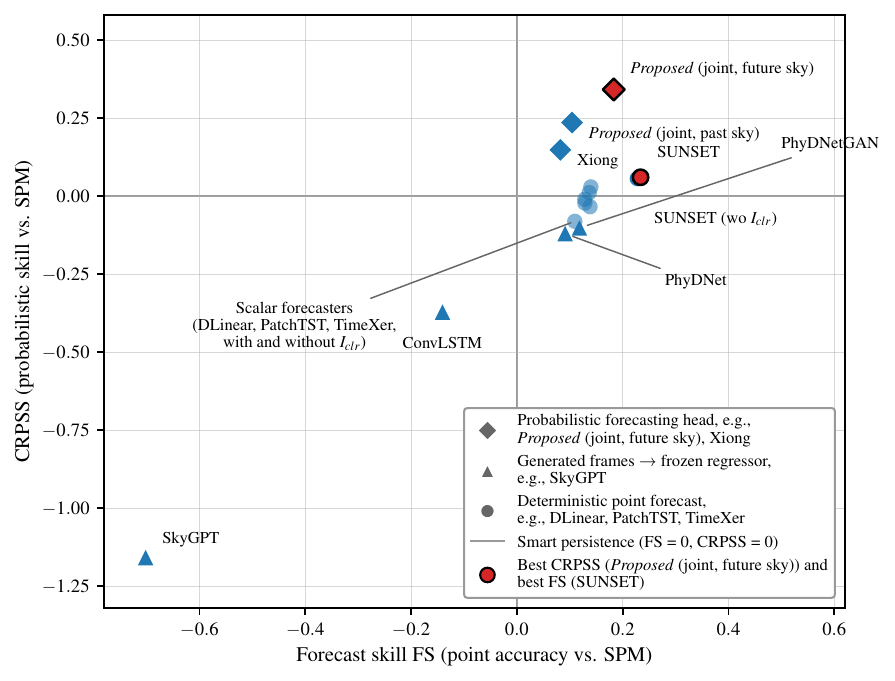}
  \caption{Forecast skill against CRPSS, every method plotted once at its headline framing.
  The two axes separate the comparison: \SunSet{} sits furthest right on point accuracy while
  \propFuture{} sits highest on probabilistic skill, and no method leads on both. Methods in the
  lower-right quadrant beat smart persistence on the mean but not on the distribution.}
  \label{fig:scatter}
\end{figure}

\paragraph{Point and probabilistic accuracy.} The proposed model attains the lowest CRPS in the
comparison, $36.9$~W\,m$^{-2}$ against $56.1$~W\,m$^{-2}$ for smart persistence, and hence the
highest CRPSS, $+0.341$, whose confidence interval lies entirely above zero. Its pooled RMSE of
$104.1$~W\,m$^{-2}$ is the second lowest; \SunSet{} attains $97.6$~W\,m$^{-2}$, with a CRPS of
$52.7$~W\,m$^{-2}$, equal to its MAE since its forecast is a single value. The two results
concern different properties of the forecast (Figure~\ref{fig:scatter}). \SunSet{} is deterministic
and provides no predictive distribution, so its advantage is confined to the point forecast, whereas
the proposed model provides a predictive distribution whose skill exceeds that of every baseline. The
point-accuracy gap to \SunSet{} is nevertheless a genuine deficit of the proposed model.

\paragraph{Scalar forecasters.} DLinear, PatchTST and TimeXer reach $109$--$113$~W\,m$^{-2}$, with
forecast skill between $+0.11$ and $+0.14$, but their CRPS, $54.5$--$60.6$~W\,m$^{-2}$, is close to
that of smart persistence, and their CRPSS values lie between $-0.08$ and $+0.03$: they improve over persistence in the mean but not in the distribution. Adding $I_{clr}$ lowers
the RMSE of all three, by $1.0$--$3.6$~W\,m$^{-2}$.

\paragraph{Video predictors.} Read through the frozen regressor, the video predictors range from
$112.3$~W\,m$^{-2}$ (\PhyDNetGAN{}) to $145.3$~W\,m$^{-2}$ (\ConvLSTM{}), and all three have a CRPS
above that of smart persistence, $61.8$--$77.0$~W\,m$^{-2}$, and hence negative CRPSS, from $-0.37$ to
$-0.10$.

\paragraph{Published generative forecasters.} \SkyGPT{} is the weakest method in the comparison,
with an RMSE of $216.8$~W\,m$^{-2}$, a forecast skill of $-0.70$, and a CRPS of
$121.1$~W\,m$^{-2}$ (CRPSS $-1.16$), i.e.\ worse than persistence on both. Our
reimplementation required substantial correction before it was a fair comparator
(Appendix~\ref{sec:app-baselines}), and Section~\ref{sec:cascade} separates how much of the gap to
the single-stage design is due to reading frames through a separate regressor rather than to frame
quality. The model of
\citet{xiong2025multimodal} has the lowest CRPS among the baselines, $47.8$~W\,m$^{-2}$ (CRPSS $+0.148$), yet remains
behind both arms of this work despite reading the irradiance history. Within this work, \propFuture{}
improves over \propPast{} by $10.0$~W\,m$^{-2}$ in RMSE and by $6.0$~W\,m$^{-2}$ in CRPS ($36.9$
against $42.9$, or $0.106$ in CRPSS); Section~\ref{sec:ablation-source}
analyses this difference.

\subsection{Cross-year performance}
\label{sec:crossyear}

The best method of each baseline group and both arms of the proposed model are retrained on the
2021--2022 data and evaluated on the same matched 2023 subset, against the same reference
(Section~\ref{sec:protocol}). The image stack of the cross-year runs was trained at a different VAE
learning rate, a stage frozen throughout joint training. Table~\ref{tab:crossyear} reports the cross-year performance --- RMSE, forecast skill and CRPSS ---
beside the in-period scores.  All six methods remain within $1.8$~W\,m$^{-2}$ in RMSE and within
$0.03$ in CRPSS of their in-period scores. The ranking is therefore unchanged from the in-period comparison. 

Under the year shift, the proposed model's forecast skill falls by $0.013$. The change is not uniform across lead times: the cross-year model is
marginally better at the shortest horizons and worse at the longest, by $0.05$ at the eighth step
(Table~\ref{tab:perhorizon}).
\SunSet{} shows the same pattern with a smaller amplitude ($-0.030$ at the eighth step, $-0.004$
pooled), which indicates that a mild long-horizon degradation under a year shift is partly a property
of the task. The CRPSS of the proposed model
moves from $+0.341$ to $+0.343$, a change of $0.002$, which is smaller than the sampler variation of
Section~\ref{sec:noise} ($0.006$).  The scores of the model of \citet{xiong2025multimodal} are equally stable.  

\begin{table}[t]
\centering
\caption{Cross-year retraining. Methods trained on a different calendar year and evaluated on the
same test set and reference as Table~\ref{tab:comparison}. $\Delta$ is cross-year minus in-period, computed before rounding;
values near zero indicate the result does not depend on which year the model was trained on. FS is
computed from the RMSE columns against the same smart-persistence reference, $127.4$~W\,m$^{-2}$. \underline{\textbf{Bold underlined}}: best in each column; \textbf{bold}: second best; for $\Delta$, closest to zero. Smart persistence is not ranked.}
\label{tab:crossyear}
\footnotesize
\setlength{\tabcolsep}{1pt}
\begin{tabular}{@{}Q{\wMethodCY}P{\wMetricCYn}P{\wMetricCYn}P{\wMetricCYn}P{\wMetricCYn}P{\wMetricCYn}P{\wMetricCYn}P{\wMetricCYn}P{\wMetricCYn}P{\wMetricCYn}@{}}
\toprule
& \multicolumn{3}{c}{RMSE\,\arrowdown{} [W\,m$^{-2}$]} & \multicolumn{3}{c}{FS\,\arrowup{}} & \multicolumn{3}{c}{CRPSS\,\arrowup{}} \\
\cmidrule(lr){2-4}\cmidrule(lr){5-7}\cmidrule(lr){8-10}
Method & in-period & cross-year & $\Delta$\,\arrowzero{} & in-period & cross-year & $\Delta$\,\arrowzero{} & in-period & cross-year & $\Delta$\,\arrowzero{} \\
\midrule
\DTLforeach*[\equal{\Grp}{0}]{cyr}{\Grp=group,\Mth=method,\RI=rin,\RC=rcy,\RD=rdel,\FI=fin,\FC=fcy,\FD=fdel,\CI=cin,\CC=ccy,\CD=cdel,\Bri=brin,\Brc=brcy,\Brd=brdel,\Bfi=bfin,\Bfc=bfcy,\Bfd=bfdel,\Bci=bcin,\Bcc=bccy,\Bcd=bcdel}{%
  \DTLiffirstrow{}{\\}\csname\Mth\endcsname & \csname\Bri\endcsname{\RI} & \csname\Brc\endcsname{\RC} & \csname\Brd\endcsname{\RD} & \csname\Bfi\endcsname{\FI} & \csname\Bfc\endcsname{\FC} & \csname\Bfd\endcsname{\FD} & \csname\Bci\endcsname{\CI} & \csname\Bcc\endcsname{\CC} & \csname\Bcd\endcsname{\CD}}\\
\midrule
\DTLforeach*[\equal{\Grp}{1}]{cyr}{\Grp=group,\Mth=method,\RI=rin,\RC=rcy,\RD=rdel,\FI=fin,\FC=fcy,\FD=fdel,\CI=cin,\CC=ccy,\CD=cdel,\Bri=brin,\Brc=brcy,\Brd=brdel,\Bfi=bfin,\Bfc=bfcy,\Bfd=bfdel,\Bci=bcin,\Bcc=bccy,\Bcd=bcdel}{%
  \DTLiffirstrow{}{\\}\csname\Mth\endcsname & \csname\Bri\endcsname{\RI} & \csname\Brc\endcsname{\RC} & \csname\Brd\endcsname{\RD} & \csname\Bfi\endcsname{\FI} & \csname\Bfc\endcsname{\FC} & \csname\Bfd\endcsname{\FD} & \csname\Bci\endcsname{\CI} & \csname\Bcc\endcsname{\CC} & \csname\Bcd\endcsname{\CD}}\\
\midrule
\DTLforeach*[\equal{\Grp}{2}]{cyr}{\Grp=group,\Mth=method,\RI=rin,\RC=rcy,\RD=rdel,\FI=fin,\FC=fcy,\FD=fdel,\CI=cin,\CC=ccy,\CD=cdel,\Bri=brin,\Brc=brcy,\Brd=brdel,\Bfi=bfin,\Bfc=bfcy,\Bfd=bfdel,\Bci=bcin,\Bcc=bccy,\Bcd=bcdel}{%
  \DTLiffirstrow{}{\\}\csname\Mth\endcsname & \csname\Bri\endcsname{\RI} & \csname\Brc\endcsname{\RC} & \csname\Brd\endcsname{\RD} & \csname\Bfi\endcsname{\FI} & \csname\Bfc\endcsname{\FC} & \csname\Bfd\endcsname{\FD} & \csname\Bci\endcsname{\CI} & \csname\Bcc\endcsname{\CC} & \csname\Bcd\endcsname{\CD}}\\
\bottomrule
\end{tabular}
\end{table}

%% file: sec_cascade.tex
\section{Effect of removing the regression stage}
\label{sec:cascade}

The proposed model is well ahead of the two-stage cascades of Table~\ref{tab:comparison} (\eg,
\SkyGPT{}). There are two possible reasons: either it generates better sky images, or its jointly
trained irradiance head maps the generated sky to irradiance better than a separate regressor does.
To separate the two, we apply the frozen U-Net regressor of Section~\ref{sec:protocol}, which plays the
role of the second stage of \SkyGPT{}~\citep{nie2024skygpt}, to the true future frames, to reference
frames and to the frames generated by the proposed model. We then compare it with the model's own
irradiance head on the same samples. RMSE follows Eq.~\eqref{eq:rmse}, with the ensemble mean as the point forecast;
single-member values are given for reference.

\begin{table}[t]
\centering
\caption{Irradiance recovered from future sky frames by two readers, on the same $1998$ samples: the
frozen U-Net regressor, applied frame by frame, and the irradiance head of the proposed model, which
reads its own denoised estimate of the future sky. ``Mean of ten members'' is the RMSE of the mean of
the ten per-member forecasts, the convention of Table~\ref{tab:comparison}; ``one member'' is the
RMSE of a single member, averaged over members. The first three rows use no generated frame. Lower is
better; bold marks the best forecast.}
\label{tab:probe}
\footnotesize
\setlength{\tabcolsep}{2pt}
\begin{tabular}{@{}Q{\wProbeWhat}P{\wProbeWho}P{\wProbeRMSE}@{}}
\toprule
Frames read & Read by & RMSE\,\arrowdown{} [W\,m$^{-2}$] \\
\midrule
\DTLforeach*{probe}{\What=what,\Who=who,\RM=rmse,\Bm=best}{%
  \DTLiffirstrow{}{\\}\What & \Who & \csname\Bm\endcsname{\RM}}\\
\bottomrule
\end{tabular}
\end{table}

Table~\ref{tab:probe} gives the results. On the true future frames, the regressor recovers irradiance
at $63.5$~W\,m$^{-2}$, nearly constant across the eight horizons ($61.3$--$65.7$,
Table~\ref{tab:perhorizon}) and well below \SunSet{} ($97.6$~W\,m$^{-2}$) and smart persistence
($127.4$~W\,m$^{-2}$): if the future sky were known, irradiance would follow from it accurately.
Passing the true frames through the model's VAE first raises the error to $80.1$~W\,m$^{-2}$, the
cost of the $4\times8\times8$ latent representation of each frame alone. On the frames generated by the proposed model
the regressor reaches $132.2$~W\,m$^{-2}$ for the mean of the ten members and $150.8$~W\,m$^{-2}$
for a single member, rising with lead time as frame quality falls (Table~\ref{tab:perhorizon}). Read
this way, the generated frames carry no more irradiance information than the last input frame
repeated over the horizon, which the same regressor reads at $127.9$~W\,m$^{-2}$ (paired difference
$+4.3$~W\,m$^{-2}$, $95\%$ interval $[-5.6,+13.8]$).

The irradiance head reads the same generated sky at $104.1$~W\,m$^{-2}$ ($134.6$ for a single
member): $28.2$~W\,m$^{-2}$ ($[12.9, 42.3]$) below the regressor reading the generated frames, and
$8.3$~W\,m$^{-2}$ ($[2.5, 14.2]$) below the best two-stage cascade, \PhyDNetGAN{}
($112.3$~W\,m$^{-2}$, Table~\ref{tab:comparison}). Read through a regressor, the generated frames are
no more informative than persistence; read by the jointly trained irradiance head, the same generated
sky gives a forecast $28$~W\,m$^{-2}$ more accurate. Of the two explanations in the opening paragraph,
the second holds: the advantage lies in the jointly trained irradiance head rather than in the frames
themselves.

To verify which information improves the irradiance head, Section~\ref{sec:ablation-source} compares
arms that differ in what it reads and in whether an image branch is trained with it, from no image
branch at all to the denoised estimate of the future sky. Section~\ref{sec:imagequality} scores the quality of the generated frames directly, and
Section~\ref{sec:ramps} shows one ramp case of generated frames and the irradiance that goes with them.

%% file: sec_imagequality.tex
\section{Quality of the generated frames}
\label{sec:imagequality}

The frozen regressor of Section~\ref{sec:cascade} is a single, indirect judge of frame quality: a deterministic estimator of this kind
penalizes a sharp cloud in the wrong place more than a blurred one. Table~\ref{tab:imagequality} therefore scores the frames of every frame-producing method directly
against the true future frames, with two metrics. The learned perceptual image patch similarity
(LPIPS, AlexNet features)~\citep{zhang2018lpips} rewards realistic texture; the peak signal-to-noise
ratio (PSNR) rewards pixel agreement and therefore favors blurred frames. PSNR is computed inside the circular sky mask of the
fisheye image ($75.9\%$ of the pixels), so the black corners do not count. For the stochastic methods each metric is given for a single member,
averaged over members, and for the pixel-wise mean of the ten members. Two reference rows bound the
table: the true frames passed through the proposed model's VAE, the best a latent-space generator of
this resolution can do, and the last input frame repeated over the horizon.

\begin{table}[t]
\centering
\caption{Quality of the generated frames on the matched test subset, pooled over the $1998$ samples
and eight horizons. LPIPS on the full frame; PSNR (dB) inside the sky mask. RMSE is the
frozen U-Net regressor applied to the same frames (W\,m$^{-2}$), so for \Xiong{} it differs from the
model's own irradiance head in Table~\ref{tab:comparison}. ``One member'' averages the metric over
the ten members; ``mean of ten'' scores the pixel-wise mean frame (for RMSE, the mean of the ten
per-member regressions). \SkyGPT{} and \Xiong{} were re-sampled for this table; their Table~\ref{tab:comparison}
scores reproduce within the bootstrap intervals. \underline{\textbf{Bold underlined}}: best forecast in
each column; \textbf{bold}: second and third best. The VAE row is a ceiling, not a forecast, and is not
ranked.}
\label{tab:imagequality}
\footnotesize
\setlength{\tabcolsep}{2pt}
\begin{tabular}{@{}Q{\wImgM}P{\wImgK}P{\wImgN}P{\wImgN}P{\wImgN}@{}}
\toprule
Frames & members & LPIPS\,\arrowdown{} & PSNR\,\arrowup{} & RMSE\,\arrowdown{} \\
\midrule
\DTLforeach*[\equal{\Grp}{1}]{img}{\Grp=grp,\Arm=arm,\Fr=frames,\Lp=lpips,\Ps=psnr,\Rm=rmse,\Bl=bl,\Bp=bp,\Br=br}{%
  \DTLiffirstrow{}{\\}\csname\Arm\endcsname & \Fr & \csname\Bl\endcsname{\Lp} & \csname\Bp\endcsname{\Ps} & \csname\Br\endcsname{\Rm}}\\
\midrule
\DTLforeach*[\equal{\Grp}{2}]{img}{\Grp=grp,\Arm=arm,\Fr=frames,\Lp=lpips,\Ps=psnr,\Rm=rmse,\Bl=bl,\Bp=bp,\Br=br}{%
  \DTLiffirstrow{}{\\}\csname\Arm\endcsname & \Fr & \csname\Bl\endcsname{\Lp} & \csname\Bp\endcsname{\Ps} & \csname\Br\endcsname{\Rm}}\\
\midrule
\DTLforeach*[\equal{\Grp}{3}]{img}{\Grp=grp,\Arm=arm,\Fr=frames,\Lp=lpips,\Ps=psnr,\Rm=rmse,\Bl=bl,\Bp=bp,\Br=br}{%
  \DTLiffirstrow{}{\\}\csname\Arm\endcsname & \Fr & \csname\Bl\endcsname{\Lp} & \csname\Bp\endcsname{\Ps} & \csname\Br\endcsname{\Rm}}\\
\bottomrule
\end{tabular}
\end{table}

Three results follow. First, among the generative forecasters the proposed model's frames are the
closest to the true sky on every metric. Paired day-block intervals for a single member exclude zero
against \Xiong{} (LPIPS $-0.0024$ $[-0.0040,-0.0007]$, PSNR $+1.03$~dB) and against
\SkyGPT{} (LPIPS $-0.0204$ $[-0.0240,-0.0168]$, PSNR $+1.61$~dB). Second, no single
image metric ranks the forecasts consistently. \PhyDNet{} and \PhyDNetGAN{} reach the highest PSNR
and the lowest regressor RMSE but the worst LPIPS: their frames converge to a smooth mean sky.
The last input frame, a real image of a sky that has since moved, has the lowest LPIPS of any
forecast ($0.043$ against $0.058$ for a member of the proposed model; paired difference $+0.0150$
$[+0.0119,+0.0180]$) but a low PSNR. LPIPS rewards realism, PSNR rewards averaging,
and neither measures whether the clouds are in the right place. Third, frame quality and forecast
skill come apart. The frames of \propFuture{} and \propPast{} are almost indistinguishable (LPIPS
$0.0577$ against $0.0591$, PSNR $+0.09$~dB), yet their irradiance heads
differ by $10.0$~W\,m$^{-2}$ (Section~\ref{sec:ablation-source}): the forecast difference lies in what
the irradiance head reads, not in the frames. The regressor nevertheless reads \propFuture{}'s frames
$14.4$~W\,m$^{-2}$ more accurately than \propPast{}'s ($[9.2, 18.5]$, mean of ten), so the two arms'
frames differ in irradiance-relevant content that the image metrics do not register. Every forecast
remains far from the VAE ceiling (LPIPS $0.006$, PSNR $31.9$~dB).

\begin{figure}[pos=!htbp]
  \centering
  \includegraphics[width=\linewidth]{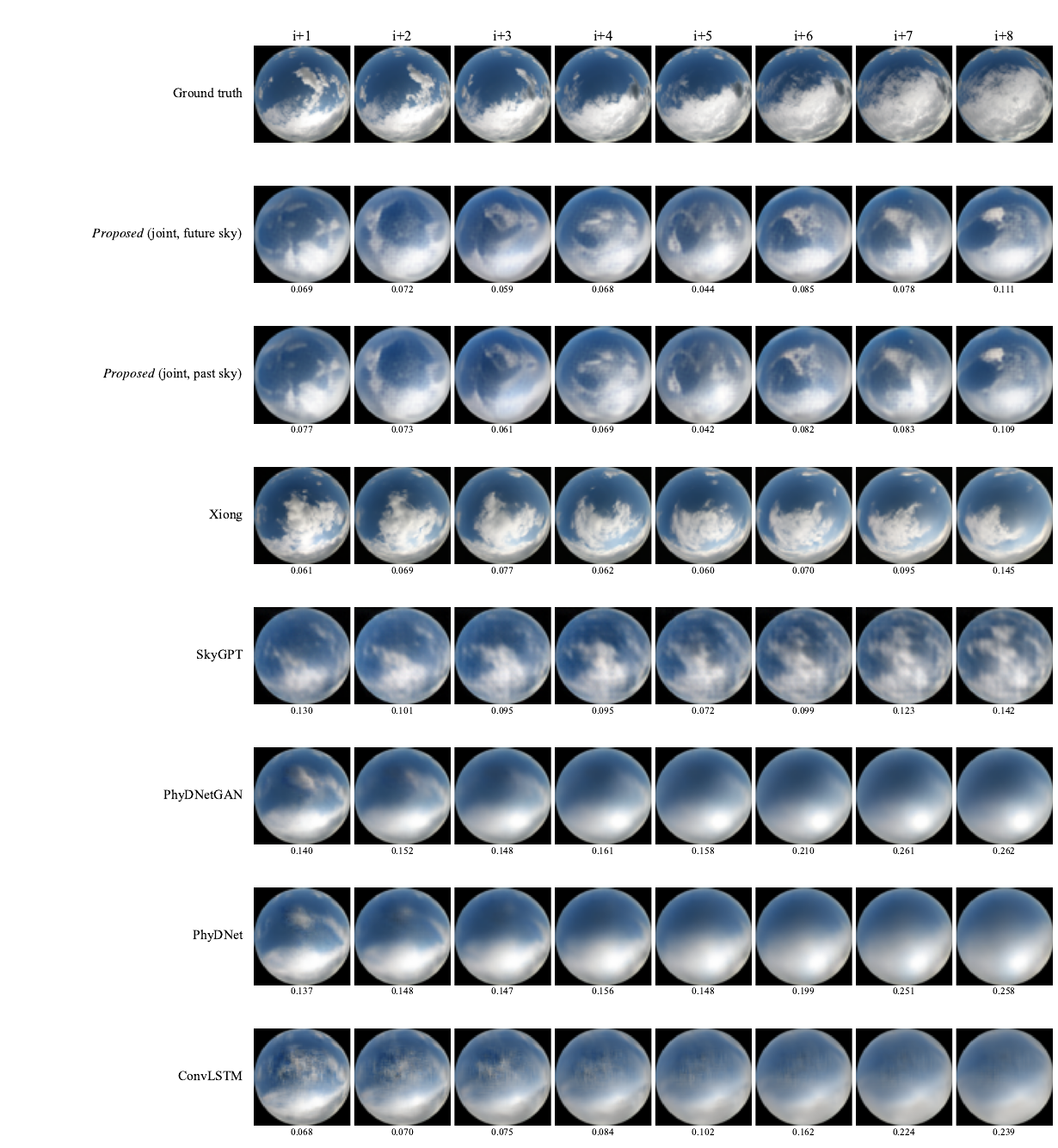}
  \caption{True future frames (top row) and generated frames of every frame-producing method for one
  partly cloudy window (2023-09-02, origin 15:28 local time, solar elevation $44^\circ$). For the
  stochastic methods the first ensemble member is shown, not the ensemble mean. The number under each
  frame is its LPIPS against the true frame. The window was selected, among partly cloudy windows of the
  middle tertile, as one where the proposed model has the lowest mean LPIPS of all rows; it illustrates
  behavior and is not representative of average performance (Table~\ref{tab:imagequality}).}
  \label{fig:examplewindow}
\end{figure}

Figure~\ref{fig:examplewindow} shows one partly cloudy window, selected as stated in its caption.
The deterministic predictors lose the cloud structure within a few steps and converge to a smooth
gradient. \SkyGPT{} is blurred, and \Xiong{} keeps sharp, realistic clouds in the wrong place. The two
arms of this work keep cloud-like texture and structure but, like every method, miss the cloud deck
that advances over the last steps.

%% file: sec_calibration.tex
\section{Per-horizon, sky-condition and ramp-event performance}
\label{sec:calibration}

This section examines performance across lead times (Section~\ref{sec:horizon}) and across sky
conditions (Section~\ref{sec:regimes}). Section~\ref{sec:reliability} then assesses the reliability of
the predictive intervals on non-ramp and ramp events and how their width responds to the size of a
ramp, and Section~\ref{sec:ramps} the accuracy on ramp events, with one example case.

\subsection{Per-horizon performance}
\label{sec:horizon}

\begin{figure}[pos=!htbp]
	\centering
	\begin{tabular}{ccc}
		\multicolumn{3}{c}{\includegraphics[width=0.9\linewidth]{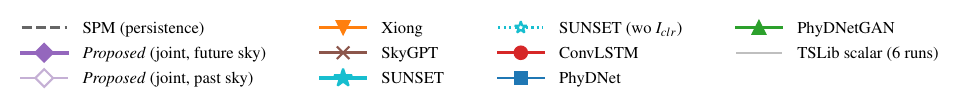}}
		\\
		\includegraphics[width=0.31\linewidth]{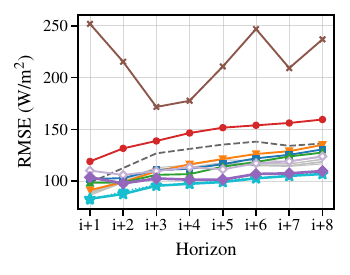} &
		\includegraphics[width=0.31\linewidth]{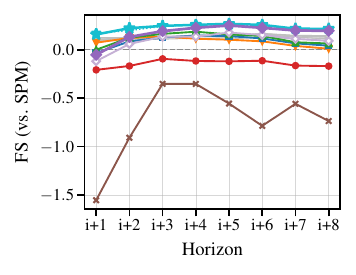} &
		\includegraphics[width=0.31\linewidth]{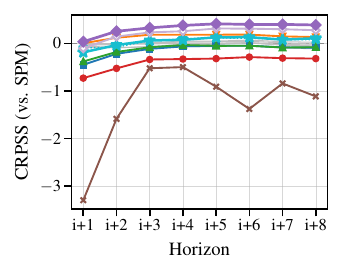}  \\
		(a) RMSE & (b) Forecast skill   & (c) CRPSS
	\end{tabular}
	\caption{Performance against lead time: (a) RMSE, (b) forecast skill and (c) CRPSS, for every method in the comparison, each at its headline framing. Forecast skill and CRPSS are computed against the same smart-persistence reference as
		Table~\ref{tab:comparison}.
		\SkyGPT{}'s chunked rollout is visible as the unstable worst curve; the scalar forecasters are
		bracketed as one muted group.} \label{fig:perhorizon} \label{fig:perhorizonfs} \label{fig:perhorizoncrpss}
\end{figure}

Performance across lead times --- (a) RMSE, (b) forecast skill and (c) CRPSS --- is shown in
Figure~\ref{fig:perhorizon}. The proposed model's forecast skill is negative at the first step and
approaches \SunSet{}'s with horizon without overtaking it at any horizon, so its pooled advantage over smart persistence in
RMSE is a mid-to-late-horizon effect. Its CRPSS is above \SunSet{}'s, and above every other
method's, at every horizon, rising from $+0.03$ at $i{+}1$ to about $+0.41$ and settling near
$+0.39$ (Table~\ref{tab:perhorizon}).

\subsection{Performance across sky conditions}
\label{sec:regimes}

Table~\ref{tab:regime} compares the proposed model's CRPSS and RMSE with those of \SunSet{}, the
strongest deterministic comparator, within each sky-condition tertile of Section~\ref{sec:protocol}
(Figure~\ref{fig:regimes}).
The proposed model has a positive CRPSS margin in all three tertiles, and the margin is close to flat; \SunSet{} keeps the better RMSE in each of them. Part of this margin reflects the difference between a probabilistic and a deterministic forecast rather than the generated sky: the irradiance head in \ArmScalar{} (Section~\ref{sec:ablation-source}) already exceeds \SunSet{}'s pooled CRPSS by $0.159$ ($[0.112, 0.210]$), and reading the denoised estimate of the future sky adds $0.114$. 

\begin{table}[h]
	\centering
	\caption{CRPSS and RMSE by sky-condition tertile on the matched test subset ($666$ samples each),
		for the proposed model and \SunSet{}, the strongest deterministic comparator. The lead is the
		proposed model's CRPSS minus \SunSet{}'s within the same tertile. RMSE in W\,m$^{-2}$. Bold marks the
		better method of each metric: the proposed model on CRPSS and \SunSet{} on RMSE, in every
		tertile.}
	\label{tab:regime}
	\footnotesize
	\setlength{\tabcolsep}{2pt}
	\begin{tabular}{@{}Q{\wRgmT}P{\wRgmN}P{\wRgmN}P{\wRgmN}P{\wRgmN}P{\wRgmN}@{}}
		\toprule
		& \multicolumn{3}{c}{CRPSS\,\arrowup{}} & \multicolumn{2}{c}{RMSE\,\arrowdown{}} \\
		\cmidrule(lr){2-4}\cmidrule(lr){5-6}
		Tertile & \propFuture{} & \SunSet{} & lead & \propFuture{} & \SunSet{} \\
		\midrule
		\DTLforeach*{rgm}{\Tr=tertile,\Cp=cprop,\Cs=csun,\Cl=clead,\Rp=rprop,\Rs=rsun,\Bcp=bcprop,\Bcs=bcsun,\Bcl=bclead,\Bp=brprop,\Bs=brsun}{%
			\DTLiffirstrow{}{\\}\Tr & \csname\Bcp\endcsname{\Cp} & \csname\Bcs\endcsname{\Cs} & \csname\Bcl\endcsname{\Cl} & \csname\Bp\endcsname{\Rp} & \csname\Bs\endcsname{\Rs}}\\
		\bottomrule
	\end{tabular}
\end{table} 

\subsection{Reliability of the predictive intervals}
\label{sec:reliability}
Table~\ref{tab:ramps} gives the method-by-method comparison on reliability of the predictive intervals. No method in this comparison reaches nominal coverage: every $\mathrm{PICP}_{90}$ is below $0.9$ off
ramps. Off ramps, two of the four probabilistic forecasters are under-dispersed, \Xiong{} is over-dispersed
and \propFuture{} is roughly calibrated; the deterministic methods have no spread and so cover no
observation.  
We therefore compare how each method's reliability changes from non-ramp to ramp events.
\Xiong{}'s non-ramp coverage of $0.847$ is higher than \propFuture{}'s $0.631$, but from non-ramp to
ramp \propFuture{}'s coverage falls by $0.026$ and \Xiong{}'s by $0.206$. The SSR column accounts for the difference. \Xiong{}'s non-ramp intervals are
over-dispersed ($\mathrm{SSR}=0.806$, below the $1.05$ target): its ensemble spread is wider than its
own forecast error, so its intervals capture observations by their width. On ramps the same quantity
rises to $1.796$, far above target, and coverage falls with it. \propFuture{} runs from
$0.972$ to $1.227$: roughly calibrated off ramps and modestly under-dispersed on them. Its intervals therefore change little between regimes, though they are neither the widest nor the
closest to nominal. \Xiong{}'s row in Table~\ref{tab:ramps} is scored at its own stored ensemble size of $100$ draws,
larger than the ten members used for every other row. Re-scoring it at ten members leaves its coverage advantage intact, so the
comparison of changes between regimes is unaffected. 

\begin{table}[h]
	\centering
	\caption{Interval coverage and spread-skill ratio (SSR, Eq.~\eqref{eq:picp}) on ramp and non-ramp events. $\Delta$ is ramp minus
		non-ramp; a small $\Delta$ means the forecast's reliability does not change when conditions become
		hard. SSR above its $1.05$ target means intervals too narrow (under-dispersed), below it too
		wide (over-dispersed); the target is derived in Section~\ref{sec:metrics}. Deterministic methods are grouped: a point
		forecast has zero coverage by construction and no spread to report, and is excluded from the
		per-column ranking. \underline{\textbf{Bold underlined}}: best in each column; \textbf{bold}: second best: highest coverage, $\Delta$ closest
		to zero, SSR closest to the $1.05$ target.}
	\label{tab:ramps}
	\footnotesize
	\setlength{\tabcolsep}{2pt}
	\begin{tabular}{@{}Q{\wRmpM}P{\wRmpC}P{\wRmpC}P{\wRmpD}P{\wRmpC}P{\wRmpC}P{\wRmpD}@{}}
		\toprule
		& \multicolumn{3}{c}{PICP90\,\arrowup{}} & \multicolumn{3}{c}{SSR ($\rightarrow 1.05$)} \\
		\cmidrule(lr){2-4}\cmidrule(lr){5-7}
		Method & non-ramp & ramp & $\Delta$\,\arrowzero{} & non-ramp & ramp & $\Delta$\,\arrowzero{} \\
		\midrule
		\DTLforeach*{rmp}{\Mth=method,\Pn=picpn,\Pr=picpr,\Pd=picpd,\Sn=ssn,\Sr=ssr,\Sd=ssd,\Bpn=bpicpn,\Bpr=bpicpr,\Bp=bpicpd,\Bsn=bssn,\Bsr=bssr,\Bs=bssd}{%
			\DTLiffirstrow{}{\\}\csname\Mth\endcsname & \csname\Bpn\endcsname{\Pn} & \csname\Bpr\endcsname{\Pr} & \csname\Bp\endcsname{\Pd} & \csname\Bsn\endcsname{\Sn} & \csname\Bsr\endcsname{\Sr} & \csname\Bs\endcsname{\Sd}}\\
		\bottomrule
	\end{tabular}
\end{table}

\begin{figure}[pos=!htbp]
	\centering
	\includegraphics[width=\linewidth]{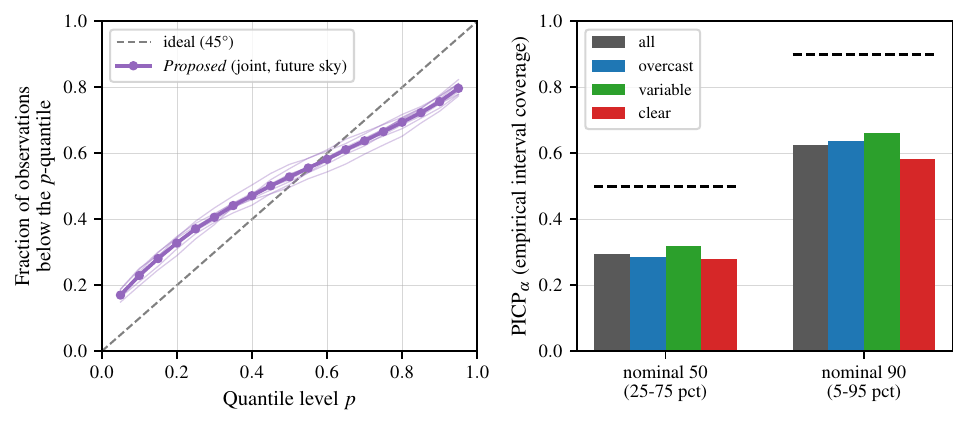}
	\caption{(a) Quantile reliability of the proposed model: for each quantile level $p$ of the ten-member
		ensemble, swept over nineteen levels from $0.05$ to $0.95$, the fraction of observations that fall
		below the ensemble's $p$-quantile; pooled (bold) and per horizon (faint). The diagonal is perfect
		reliability; a curve above it at low $p$ and below it at high $p$ means the ensemble's quantiles are
		not extreme enough. (b) $\mathrm{PICP}_\alpha$ at $\alpha=0.5$ and $0.9$ by
		sky-condition tertile; dashed lines are the nominal targets.}
	\label{fig:reliability}
\end{figure}

Figure~\ref{fig:reliability}(a) examines the proposed model's quantiles. Over nineteen quantile levels
$p$ from $0.05$ to $0.95$, the fraction of observations below the ensemble's $p$-quantile crosses the
diagonal at $p\approx0.56$, close to the median, rather than lying to one side of it. It exceeds $p$ by
$0.127$ on average in the low tail ($p\le0.15$) and falls short by $0.141$ in the high tail
($p\ge0.85$). The forecast is thus nearly centered, with quantiles that are not extreme enough: the
signature of under-dispersion rather than of a mis-centered forecast. $\mathrm{PICP}_{90}$ is flat across horizons ($0.600$--$0.662$, Table~\ref{tab:perhorizon}), with no degradation
at longer lead times.

\begin{figure}[pos=!htbp]
  \centering
  \begin{tabular}{cc}
    \includegraphics[width=0.48\linewidth]{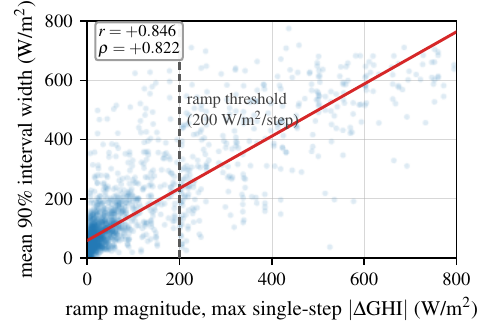} &
    \includegraphics[width=0.48\linewidth]{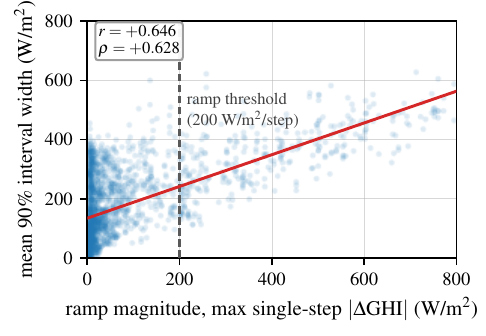} \\
    (a) \propFuture{} & (b) \Xiong{}
  \end{tabular}
  \caption{Width of the $90\%$ prediction interval against ramp magnitude, one point per test sample,
  for (a) \propFuture{} and (b) \Xiong{}, on the same axes. Width is the $5$--$95$ percentile
  spread of the ensemble, averaged over the eight horizons; magnitude is the largest single-step
  $|\Delta\mathrm{GHI}|$ in the forecast window. The dashed line is the $200$~W\,m$^{-2}$ ramp
  threshold: the ramp subset is the region to its right. \Xiong{} is subsampled to ten members so the
  two panels are comparable. Red lines are least-squares fits, of slope $0.88$ and $0.54$.}
  \label{fig:widthramp}
\end{figure}

\paragraph{What the intervals give an operator.} A deterministic forecast offers no width to act on:
\SunSet{} and the scalar forecasters return a single value in both regimes. The proposed model's
$90\%$ interval instead widens with the event, from a mean of $87$~W\,m$^{-2}$ off ramps to
$417$~W\,m$^{-2}$ on them, and its width tracks the size of the event rather than merely its presence:
correlated with the per-sample ramp magnitude, the mean width gives $r=+0.85$ (Spearman
$\rho=+0.82$), stable across all eight horizons. \Xiong{} and \SkyGPT{} widen less and track magnitude
more loosely (Table~\ref{tab:widthramp} and Figure~\ref{fig:widthramp}), and \Xiong{}'s interval
opens by $0.54$~W\,m$^{-2}$ per W\,m$^{-2}$ of ramp against the proposed model's $0.88$.

\begin{table}[t]
\centering
\caption{Response of the $90\%$ prediction interval to ramp magnitude, on the matched test subset.
$r$ and $\rho$ are the Pearson and Spearman correlations between a sample's mean interval width and
its ramp magnitude (largest single-step $|\Delta\mathrm{GHI}|$); widths are means off and on the ramp
subset, in W\,m$^{-2}$, and the ratio is their quotient. All methods at $M=10$ members, \Xiong{}
subsampled to match. Deterministic methods are absent: a point forecast has no width.
\underline{\textbf{Bold underlined}}: best in each correlation column; \textbf{bold}: second best. Widths and
their ratio describe the interval rather than score it and are not ranked.}
\label{tab:widthramp}
\footnotesize
\setlength{\tabcolsep}{2pt}
\begin{tabular}{@{}Q{\wWrpM}P{\wWrpN}P{\wWrpN}P{\wWrpN}P{\wWrpN}P{\wWrpN}@{}}
\toprule
Method & $r$\,\arrowup{} & $\rho$\,\arrowup{} & non-ramp & ramp & ratio \\
\midrule
\DTLforeach*{wrp}{\Mth=method,\Rr=r,\Rh=rho,\Wn=wnon,\Wr=wramp,\Rt=ratio,\Br=br,\Brh=brh}{%
  \DTLiffirstrow{}{\\}\csname\Mth\endcsname & \csname\Br\endcsname{\Rr} & \csname\Brh\endcsname{\Rh} & \Wn & \Wr & \Rt}\\
\bottomrule
\end{tabular}
\end{table}

A reserve sized on this interval therefore grows with the size of the anticipated ramp,
which is the quantity a scheduler needs and a point forecast cannot supply. The level at which it
scales must be read with the coverage of Table~\ref{tab:ramps} in mind: it is $0.605$ at a nominal
$0.90$ on ramps, so the interval ranks and scales risk correctly but does not deliver its nominal
level, and an operator would size against the empirical coverage rather than the nominal one.

\subsection{Ramp events}
\label{sec:ramps}

Ramps are defined from the irradiance trajectory rather than from the sky condition at the forecast
origin, and the two are close to orthogonal: ramps do not concentrate in
the overcast or variable tertiles. Table~\ref{tab:rampscores} scores every method on the $334$ ramp samples. In RMSE, \SunSet{} is the
most accurate ($216.4$~W\,m$^{-2}$), ahead of \propFuture{} ($226.4$). In probabilistic skill, the proposed model attains a CRPSS of $+0.440$ on the ramp subset,
against $+0.342$ for \propPast{}, $+0.310$ for \Xiong{} and at most $+0.198$ for every other method.

\begin{table}[t]
	\centering
	\caption{Forecast accuracy on the ramp subset ($n=334$ of the matched test subset), scored against the
		same smart-persistence reference as Table~\ref{tab:comparison}. RMSE in W\,m$^{-2}$; FS and CRPSS are
		skill scores, higher is better. Rows are grouped as in Table~\ref{tab:comparison}. Errors on ramps are
		about twice their pooled values for every method, so these numbers are not comparable with
		Table~\ref{tab:comparison} and none is quoted beside one. \underline{\textbf{Bold underlined}}: best in each column; \textbf{bold}: second best.}
	\label{tab:rampscores}
	\footnotesize
	\setlength{\tabcolsep}{2pt}
	\begin{tabular}{@{}Q{\wRmpMethod}P{\wRmpNum}P{\wRmpNum}P{\wRmpNum}@{}}
		\toprule
		Method & RMSE\,\arrowdown{} & FS\,\arrowup{} & CRPSS\,\arrowup{} \\
		\midrule
		\DTLforeach*[\equal{\Grp}{1}]{rsc}{\Grp=group,\Mth=method,\RM=rmse,\FS=fs,\CS=crpss,\Brm=brmse,\Bfs=bfs,\Bcs=bcrpss}{%
			\DTLiffirstrow{}{\\}\csname\Mth\endcsname & \csname\Brm\endcsname{\RM} & \csname\Bfs\endcsname{\FS} & \csname\Bcs\endcsname{\CS}}\\
		\midrule
		\DTLforeach*[\equal{\Grp}{2}]{rsc}{\Grp=group,\Mth=method,\RM=rmse,\FS=fs,\CS=crpss,\Brm=brmse,\Bfs=bfs,\Bcs=bcrpss}{%
			\DTLiffirstrow{}{\\}\csname\Mth\endcsname & \csname\Brm\endcsname{\RM} & \csname\Bfs\endcsname{\FS} & \csname\Bcs\endcsname{\CS}}\\
		\midrule
		\DTLforeach*[\equal{\Grp}{3}]{rsc}{\Grp=group,\Mth=method,\RM=rmse,\FS=fs,\CS=crpss,\Brm=brmse,\Bfs=bfs,\Bcs=bcrpss}{%
			\DTLiffirstrow{}{\\}\csname\Mth\endcsname & \csname\Brm\endcsname{\RM} & \csname\Bfs\endcsname{\FS} & \csname\Bcs\endcsname{\CS}}\\
		\midrule
		\DTLforeach*[\equal{\Grp}{4}]{rsc}{\Grp=group,\Mth=method,\RM=rmse,\FS=fs,\CS=crpss,\Brm=brmse,\Bfs=bfs,\Bcs=bcrpss}{%
			\DTLiffirstrow{}{\\}\csname\Mth\endcsname & \csname\Brm\endcsname{\RM} & \csname\Bfs\endcsname{\FS} & \csname\Bcs\endcsname{\CS}}\\
		\midrule
		\DTLforeach*[\equal{\Grp}{5}]{rsc}{\Grp=group,\Mth=method,\RM=rmse,\FS=fs,\CS=crpss,\Brm=brmse,\Bfs=bfs,\Bcs=bcrpss}{%
			\DTLiffirstrow{}{\\}\csname\Mth\endcsname & \csname\Brm\endcsname{\RM} & \csname\Bfs\endcsname{\FS} & \csname\Bcs\endcsname{\CS}}\\
		\midrule
		\DTLforeach*[\equal{\Grp}{6}]{rsc}{\Grp=group,\Mth=method,\RM=rmse,\FS=fs,\CS=crpss,\Brm=brmse,\Bfs=bfs,\Bcs=bcrpss}{%
			\DTLiffirstrow{}{\\}\csname\Mth\endcsname & \csname\Brm\endcsname{\RM} & \csname\Bfs\endcsname{\FS} & \csname\Bcs\endcsname{\CS}}\\
		\bottomrule
	\end{tabular}
\end{table}

\begin{figure}[pos=!htbp]
  \centering
  \includegraphics[width=\linewidth]{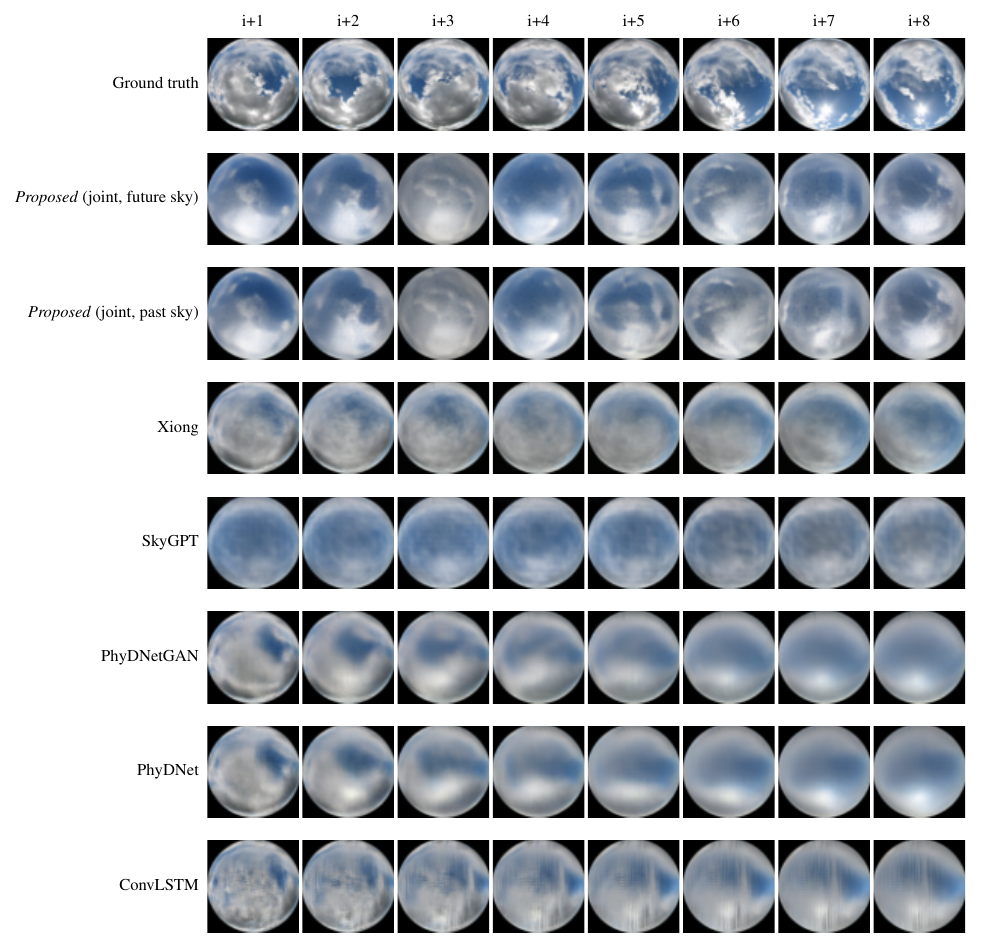}
  \caption{Ground-truth future frames (top row) and the frames generated by each method across the
  forecast horizon, for one ramp-up case. No method reproduces the ground
  truth closely; the relevant comparison is between the generated rows. Frame quality decreases with
  lead time, consistent with the per-horizon degradation of the frozen regressor reported in
  Section~\ref{sec:cascade}.}
  \label{fig:frames}
\end{figure}

\begin{figure}[pos=!htbp]
  \centering
  \includegraphics[width=0.75\linewidth]{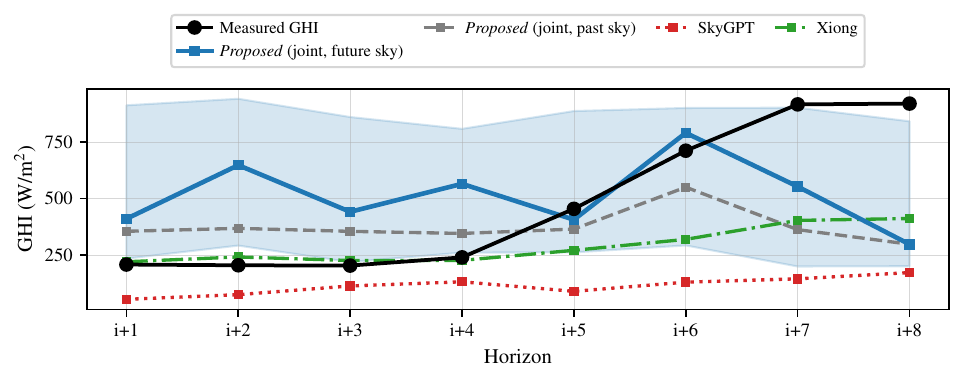}
  \caption{Measured and forecast irradiance for the same case as Figure~\ref{fig:frames}. The shaded
  band is the $5$--$95\%$ interval of the proposed model's ten-member ensemble. The measured series
  rises by about $700$~W\,m$^{-2}$ between $i{+}4$ and $i{+}7$; the deterministic rows follow it only
  partly, and the ensemble band is wide enough to contain most of the rise.}
  \label{fig:frameseries}
\end{figure}

Figures~\ref{fig:frames} and~\ref{fig:frameseries} show one ramp-up case from the test subset: the
frames each method generates across the horizon, and the irradiance that goes with them. No
method reproduces the ground truth closely, and frame quality falls with lead time, as the frozen
regressor's per-horizon errors indicate over the whole subset (Section~\ref{sec:cascade}). In
irradiance, the deterministic rows follow the $700$~W\,m$^{-2}$ rise only partly, while the proposed
model's ensemble band is wide enough to contain most of it, the case-level counterpart of the interval response in Table~\ref{tab:widthramp}.

%% file: sec_ablations.tex
\section{Ablations}
\label{sec:ablations}

Two questions remain from the comparison. The first concerns the design: which input of the
irradiance head produces the gain (Section~\ref{sec:ablation-source}). The second concerns the
remaining point-accuracy gap to \SunSet{}, which reads the measured irradiance history that the
proposed model does not receive (Section~\ref{sec:inputs}).

\subsection{Input of the irradiance head}
\label{sec:ablation-source}

Table~\ref{tab:source} compares the five arms defined in the Introduction
(Eq.~\eqref{eq:axis}; Appendix~\ref{sec:lattice}) on the matched subset, under the protocol given in
its caption. The past frames read by \ArmScalar{} and \textit{past sky} carry no information about the
future. The noised future latents read by \ArmNoisy{} and \ArmConcat{} carry it only weakly at high
noise levels. The denoised estimate read by \textit{future sky} is both clean and about the future. The
oracle row keeps the trained weights of \propFuture{} and, at evaluation time only, gives its irradiance
head the denoised estimate formed from the \emph{true} future frames, noised to each step's level as in
training. It is not a forecast; it measures how much perfect frames are worth to the trained irradiance
head.

\begin{table}[t]
\centering
\caption{What the irradiance head reads and whether an image branch is trained with it, on the matched test subset ($50$ sampling steps, $M=10$, one
sampler draw per test sample for every row, so the rows share their initial noise). Each arm is scored
at its own best-validation checkpoint. SSR target $1.05$ (Section~\ref{sec:metrics}). The oracle row
reads the true future frames; it is not a forecast and is not ranked. \underline{\textbf{Bold
underlined}}: best in each column; \textbf{bold}: second best (SSR: closest to $1.05$).}
\label{tab:source}
\footnotesize
\setlength{\tabcolsep}{2pt}
\begin{tabular}{@{}Q{\wSrcBr}Q{\wSrcArm}P{\wSrcN}P{\wSrcN}P{\wSrcN}P{\wSrcN}P{\wSrcN}@{}}
\toprule
Image branch & Irradiance head reads & RMSE\,\arrowdown{} & FS\,\arrowup{} & CRPSS\,\arrowup{} & SSR\,($\rightarrow1.05$) & PICP90\,\arrowup{} \\
\midrule
\DTLforeach*{src}{\Brn=branch,\Arm=arm,\RM=rmse,\FS=fs,\CS=crpss,\Ss=ssr,\Pc=picp,\Br=brmse,\Bf=bfs,\Bc=bcrpss,\Bs=bssr,\Bp=bpicp}{%
  \DTLiffirstrow{}{\\}\Brn & \Arm & \csname\Br\endcsname{\RM} & \csname\Bf\endcsname{\FS} & \csname\Bc\endcsname{\CS} & \csname\Bs\endcsname{\Ss} & \csname\Bp\endcsname{\Pc}}\\
\bottomrule
\end{tabular}
\end{table}

Only the denoised estimate of the future sky improves over \ArmScalar{} in RMSE: by
$9.3$~W\,m$^{-2}$ ($[6.1, 12.0]$), with CRPSS higher by $0.114$ ($[0.086, 0.151]$). The past frames,
the noised future frames and their combination stay within $1$~W\,m$^{-2}$ of \ArmScalar{}
($113.8$--$114.7$~W\,m$^{-2}$; no paired difference among these four arms excludes zero) and
$9.4$--$10.1$~W\,m$^{-2}$ behind \textit{future sky}, with intervals that exclude zero. The future sky
is therefore useful to the irradiance head only in denoised form. Under the noise of most reverse
steps, the irradiance head extracts nothing from the noised iterate, alone or with the past frames, and
generating images without reading their denoised estimate adds nothing. The gain holds in every
sky-condition tertile and is largest on ramps, where RMSE falls from $244.3$ to $227.6$~W\,m$^{-2}$.

\ArmNoisy{} reaches the same RMSE by a different route. Its ensemble is much wider (SSR $0.98$
against $1.18$ for \propFuture{}; $\mathrm{PICP}_{90}$ $0.68$), which helps on ramps (RMSE
$229.6$~W\,m$^{-2}$, CRPSS $+0.440$) and costs off them (non-ramp MAE $47.2$ against $31.4$~W\,m$^{-2}$): the
noise of its input appears to widen the ensemble rather than inform its mean.

Given the true future frames in their training form, the same irradiance head improves by a further
$3.0$~W\,m$^{-2}$ ($[2.0, 4.2]$), so the model's own estimate already delivers about three-quarters of
the $12.2$~W\,m$^{-2}$ that perfect frames are worth. An oracle that reads the clean true frames,
unlike anything seen in training, gains no more (Appendix~\ref{sec:app-designspace}); the small
remaining gain is a property of the trained irradiance head, not of a mismatch between real and
generated inputs. Real future frames hold more than the irradiance head uses: the frozen regressor
reads them at $63.5$~W\,m$^{-2}$ (Table~\ref{tab:probe}). How much of that an irradiance head trained
on true frames could recover is not measured here.

The comparison with \propPast{} on the Table~\ref{tab:comparison} run agrees. With the same
architecture and parameter count, reading the denoised estimate instead of the past frames lowers RMSE
by $10.0$~W\,m$^{-2}$ and raises CRPSS from $+0.235$ to $+0.341$. At seed~1 \propPast{} stopped five
epochs short of \propFuture{}, all in the low-rate second phase (Appendix~\ref{sec:app-protocol}),
which can only inflate the difference; its other two seeds ran the full budget and still select epoch
$954$, so the missing epochs are unlikely to carry the gap. The difference has the same sign at all
three seeds, and its paired interval excludes zero. \ArmScalar{}, \ArmNoisy{} and \ArmConcat{} exist at
one training seed, and \ArmScalar{} trains in a single phase (Appendix~\ref{sec:app-protocol}).

Feedback in the other direction does not help. Feeding the denoised estimate of the future CSI,
$\hat{\vect{y}}_0$, back into the image branch (Appendix~\ref{sec:app-designspace}) raises RMSE by
$4.5$~W\,m$^{-2}$ ($[0.7, 7.5]$) at matched training budget, more than the noise floors of
Section~\ref{sec:noise}, with no difference in CRPSS. The other two cross-modal comparisons differ by
no more than the largest of those floors (Table~\ref{tab:pairs}). The gain therefore comes from what the
irradiance head reads, not from feedback into the image branch.

\subsection{Measured irradiance history}
\label{sec:inputs}

All configurations so far condition on past sky images and deterministic covariates only. Two results
suggest that the remaining point-accuracy gap is due to this input rather than to the architecture.
True future frames improve the irradiance head by only about $3$~W\,m$^{-2}$
(Section~\ref{sec:ablation-source}), and the method with the lowest RMSE, \SunSet{}, reads the past
irradiance series. This section tests that explanation with an ablation arm, not the proposed
configuration: the irradiance head additionally reads the eight past values of measured irradiance as
a third token stream. Each of two seeds is compared with a control trained on the same modified code
path with the stream disabled (Appendix~\ref{sec:app-protocol} describes the arm and the control).

\begin{table}[t]
\centering
\caption{Conditioning the irradiance head on eight past timesteps of measured irradiance, against a
matched control on the same code path. Both arms of each pair are scored at the ceiling epoch, $959$ (Appendix~\ref{sec:app-protocol}). Gain is control minus stream, so positive favors the stream.}
\label{tab:inputs}
\footnotesize
\setlength{\tabcolsep}{2pt}
\begin{tabular}{@{}Q{\wInpSeed}P{\wInpCtl}P{\wInpStr}P{\wInpGain}P{\wInpCrpss}@{}}
\toprule
Seed & control & $+$irradiance & RMSE gain\,\arrowup{} & CRPSS gain\,\arrowup{} \\
\midrule
\DTLforeach*{inp}{\Sd=seed,\Ct=control,\St=stream,\Gn=gain,\Cg=crpssgain}{%
  \DTLiffirstrow{}{\\}\Sd & \Ct & \St & \Gn & \Cg}\\
\bottomrule
\end{tabular}
\end{table}

\paragraph{Result.} The measured irradiance history lowers RMSE by $4.8$--$7.1$~W\,m$^{-2}$ and
raises CRPSS by about $0.09$ at both seeds (Table~\ref{tab:inputs}), a CRPS reduction of
$13.3\%$ and $12.3\%$. To the same irradiance head it is worth more than the $3.0$~W\,m$^{-2}$ gained from true
future frames, so the remaining point-accuracy gap is better addressed through the inputs than through
the quality of the generated frames. The study deliberately evaluates what a joint sky-image model achieves from the sky
alone, which keeps its comparison with the two-stage cascades of Section~\ref{sec:cascade} symmetric in inputs; extending
the model with measured irradiance is a direction whose value this ablation quantifies.

%% file: sec_conclusion.tex
\section{Conclusion}
\label{sec:conclusion}

Sky-image irradiance forecasting has largely been built as a cascade: predict the future sky, then
regress irradiance from it. We evaluated a single-stage alternative that co-denoises the future frames
and the future clear-sky index in one diffusion process and reads the forecast from the irradiance
head. On the same samples, the irradiance head is $8.3$~W\,m$^{-2}$ more accurate than the best
two-stage cascade and $28$~W\,m$^{-2}$ more accurate than a separately trained regressor reading the
model's own frames. Part of that margin comes from the irradiance head itself: with no image branch
(\ArmScalar{}) it already reaches $113.8$~W\,m$^{-2}$. Reading the denoised estimate of the future sky
lowers RMSE by a further $9.3$~W\,m$^{-2}$, reading the past frames adds nothing, and perfect future
frames add only $3.0$~W\,m$^{-2}$ more. The generated frames are the closest to the observed sky among
the generative forecasters. Yet, read through a regressor trained on real images, they carry no more
irradiance information than the last observed frame; their value lies in the denoised estimate that
the irradiance head reads.

Four limitations bound these results. \SunSet{}, which also reads the measured irradiance series, gives
the best point forecast. The proposed model's intervals rank and scale risk but fall short of nominal
coverage. The cross-modal result rests on one feedback mechanism and one training seed per variant.
The evaluation covers one site and one test year, at $64\times64$ pixels and a sixteen-minute horizon,
and forecasts GHI rather than PV power.

Two further results shape the outlook. Feeding the future-CSI estimate back into the image branch
brought no gain and, at matched training budget, lowered point accuracy, so the advantage comes from
the denoised estimate rather than from cross-modal information flow. The point-accuracy gap to the
strongest deterministic baseline is an input difference rather than an architectural ceiling: a
perfect future sky is worth about $3$~W\,m$^{-2}$ to the trained irradiance head, eight steps of
measured irradiance $4.8$--$7.1$. A single-stage model that also reads the measured irradiance should
therefore narrow that gap while keeping the predictive distribution; the arm of
Section~\ref{sec:inputs} already comes within $3.9$--$5.8$~W\,m$^{-2}$ of \SunSet{}. The model
reported here forecasts from imagery alone, as a site with a camera and no pyranometer requires, and
its intervals widen with the size of the ramp they anticipate.

%% file: sec_appendix.tex
\appendix
\renewcommand{\theequation}{\thesection.\arabic{equation}}
\renewcommand{\thetable}{\thesection.\arabic{table}}
\renewcommand{\thefigure}{\thesection.\arabic{figure}}
\makeatletter
\@addtoreset{equation}{section}\@addtoreset{table}{section}\@addtoreset{figure}{section}
\makeatother
\setcounter{equation}{0}\setcounter{table}{0}\setcounter{figure}{0}

\section{Baseline fidelity}
\label{sec:app-baselines}

\SkyGPT{}, \ConvLSTM{}, \PhyDNet{}, \PhyDNetGAN{} and \SunSet{} were taken from the reference
implementation released with \citet{nie2024skygpt}~\citep{skygptrepo}; \DLinearIclr{}, \PatchTSTIclr{}
and \TimeXerIclr{} were taken from the Time-Series-Library~\citep{tslib}. Each was audited for
convergence and training-budget parity before being scored. We record what had to be corrected,
because in two cases the correction made our own result look worse.

\paragraph{\SunSet{} --- retrained after an early-stopping audit.} Both variants had trained past their
validation minimum, because the early-stopping patience was applied on a coarser schedule than
intended. Both were retrained with only the validation cadence changed, and both then stopped at the
intended margin after their own validation minimum. The corrected variant reaches
$97.6$~W\,m$^{-2}$, the value that beats the proposed model on pooled RMSE in
Table~\ref{tab:comparison}; the correction widened our own point-accuracy concession.

\paragraph{\SkyGPT{} --- a rollout discontinuity introduced by our port, not by the method.} Our
initial port generated the eight-frame horizon in two four-frame chunks, the second conditioned on the
first chunk's generated frames, because of a clip-length mismatch that the original implementation
avoids by generating in one pass. This produced a sharp discontinuity between the fourth and fifth lead
times, an artifact of our port rather than of the architecture. With the clip length raised so that
one sampling call covers the whole horizon, both stages were retrained and converged by patience
exhaustion like the other baselines. Table~\ref{tab:comparison} reports the corrected version.

\paragraph{\ConvLSTM{}, \PhyDNet{} and \PhyDNetGAN{} --- converged, and scored through a regressor.}
All three share the same early-stopping criteria and converged rather than being cut short. None
forecasts irradiance directly, so each is scored by passing its generated frames to the frozen U-Net
of Section~\ref{sec:cascade}: the two-stage arrangement that section measures.

\paragraph{\PhyDNetGAN{} --- no adversarial term reaches the generator.} In our implementation the
generator of \PhyDNetGAN{} is trained with the \PhyDNet{} loss alone. The frame discriminator's loss,
$0.01\,[(D(\text{true}) - 1)^2 + D(\text{generated})^2]$, is back-propagated through the generated
frames into the generator, which therefore receives the discriminator's own gradient instead of an
adversarial one. The row behaves as \PhyDNet{} with a weak additional term
(Tables~\ref{tab:comparison} and~\ref{tab:imagequality}) and should be read as a \PhyDNet{}-type
predictor. We have not retrained it with an adversarial generator loss.

\paragraph{\Xiong{} --- faithful, with a diagnosed mechanism.} The full three-stage pipeline was run
as published. Its stated source of ensemble diversity is a neural stochastic differential equation,
but the trained SDE has collapsed to near-deterministic behavior: the spread comes from the sampler
and a fitted observation-noise term. This is a diagnosis of the reimplementation's behavior, not a
criticism of the method, and it does not change the numbers we quote.

\section{Cross-modal feedback and the clean-frame oracle}
\label{sec:app-designspace}

Section~\ref{sec:ablation-source} compares what the irradiance head reads. This appendix adds a path
in the opposite direction, from the irradiance head back into the image branch, and a second version
of the oracle, both scored under the protocol of Table~\ref{tab:source}.

\begin{table}[h]
\centering
\caption{Cross-modal feedback, on the matched test subset under the protocol of
Table~\ref{tab:source}. Rows above the rule read the denoised estimate of the future sky, rows below
the past frames; each arm with the path sits below its partner without it. \ArmFiLMmatched{} and
\ArmFiLM{} are two trainings of the same variant (Table~\ref{tab:pairs}). \underline{\textbf{Bold
underlined}}: best in each column; \textbf{bold}: second best (SSR: closest to $1.05$; ties at the
printed precision share a rank).}
\label{tab:designspace}
\footnotesize
\setlength{\tabcolsep}{2pt}
\begin{tabular}{@{}Q{\wDsArm}P{\wDsN}P{\wDsN}P{\wDsN}P{\wDsN}@{}}
\toprule
Arm & RMSE\,\arrowdown{} & FS\,\arrowup{} & CRPSS\,\arrowup{} & SSR\,($\rightarrow1.05$) \\
\midrule
\DTLforeach*[\equal{\Grp}{1}]{dsp}{\Grp=group,\Arm=arm,\RM=rmse,\FS=fs,\CS=crpss,\Sp=ssr,\Br=brmse,\Bf=bfs,\Bc=bcrpss,\Bs=bssr}{%
  \DTLiffirstrow{}{\\}\csname\Arm\endcsname & \csname\Br\endcsname{\RM} & \csname\Bf\endcsname{\FS} & \csname\Bc\endcsname{\CS} & \csname\Bs\endcsname{\Sp}}\\
\midrule
\DTLforeach*[\equal{\Grp}{2}]{dsp}{\Grp=group,\Arm=arm,\RM=rmse,\FS=fs,\CS=crpss,\Sp=ssr,\Br=brmse,\Bf=bfs,\Bc=bcrpss,\Bs=bssr}{%
  \DTLiffirstrow{}{\\}\csname\Arm\endcsname & \csname\Br\endcsname{\RM} & \csname\Bf\endcsname{\FS} & \csname\Bc\endcsname{\CS} & \csname\Bs\endcsname{\Sp}}\\
\bottomrule
\end{tabular}
\end{table}

\paragraph{The oracle arm.} Neither oracle is separately trained: both use the weights of
\propFuture{} and change only what the prediction pass reads at evaluation time. The clean-frame
oracle reads the tokens of the true future frames, encoded once and used at every reverse step,
instead of those of $\hat{\vect{x}}_0$. In training, however, the irradiance head never read clean
future frames. It read $\hat{\vect{x}}_0$ formed from the true future latents noised to the sampled
step, close to the true frames at low noise and blurred at high noise; the oracle of
Section~\ref{sec:ablation-source} reproduces that input at every reverse step. Under the same sampler
seeding the clean-frame oracle scores $102.06$~W\,m$^{-2}$ and the training-matched one $101.61$, a
difference of $0.45$~W\,m$^{-2}$ ($[-3.2, +4.5]$); the latter improves over the deployed arm by
$3.0$~W\,m$^{-2}$. The small gain is therefore not a mismatch between real frames and the training
input: the trained irradiance head extracts little more from a perfect future sky than from its own
estimate. The frames carry much more, as the frozen regressor's $63.5$~W\,m$^{-2}$ on the same frames
shows; \citet{varaschin2025benchmarking} report $44.1$~W\,m$^{-2}$ for a dedicated estimator on real
frames, though their setting differs (year, split and horizons). Retraining the irradiance head on
true future frames, which would measure how much of that information this design can use, is left to
future work. The training-matched oracle's SSR, $1.13$, is closer to the $1.05$ target than the
deployed arm's $1.18$, but its $\mathrm{PICP}_{90}$ is still $0.64$ (Table~\ref{tab:source}).
Removing the uncertainty about the future sky therefore does not remove the under-coverage of
Section~\ref{sec:reliability}.

\paragraph{How the cross-modal path works.} The proposed design passes information from the image
branch to the irradiance head through $\hat{\vect{x}}_0$ (Section~\ref{sec:method-step}); the variants
below add the reverse path. Applying Eq.~\eqref{eq:x0hat} to the CSI stream with the noise estimate of
the pass that reads the clean tokens ($\epsilon'_y$ of Eq.~\eqref{eq:condpass}, or that pass's own
estimate in a single-pass variant) gives a denoised estimate of the future CSI,
$\hat{\vect{y}}_0\in\mathbb{R}^{8}$,

\begin{equation}
\hat{\vect{y}}_0 = \frac{\vect{y}_t - \sqrt{1-\bar\alpha_t}\;\mathrm{sg}[\epsilon'_y]}{\sqrt{\bar\alpha_t}},
\label{eq:y0hat}
\end{equation}
which is available before the image branch runs. The variant passes $\hat{\vect{y}}_0$ to the image
branch through the timestep embedding, a route separate from cross-attention and ControlNet. Every
UNet, including the proposed model's, maps $t$ to an embedding $\gamma(t)\in\mathbb{R}^{512}$, and each
residual block adds a learned projection of it to its feature maps as a per-channel shift. The variant
adds to $G_\theta$ a small network $\psi:\mathbb{R}\to\mathbb{R}^{512}$ and, for each future frame
$j=i+1,\dots,i+8$, replaces $\gamma(t)$ by

\begin{equation}
\tilde{\gamma}^j(t) = \gamma(t) + \psi\big(\hat{y}^j_0\big),
\label{eq:csiemb}
\end{equation}
where $\psi(u) = W_2\,\mathrm{SiLU}\big(W_1\,\phi(u)+b_1\big)+b_2$, with Fourier features
$\phi(u)=\big[u,\ \sin(2^k\pi u),\ \cos(2^k\pi u)\big]_{k=0}^{15}\in\mathbb{R}^{33}$ of the
standardized CSI value, and $\hat{y}^j_0$ is the $j$th entry of $\hat{\vect{y}}_0$. Nothing else in
$G_\theta$ changes, and we write the variant's image branch as

\begin{equation}
\epsilon_x = G_\theta\big(\vect{x}_t,\,t;\ [\,\vect{e},\,s_t\,],\ \vect{a}^i,\ \hat{\vect{y}}_0\big),
\label{eq:imgbranchcsi}
\end{equation}
with the weights of $\psi$ counted in $\theta$. This is how class-conditional diffusion models supply
a class label~\citep{dhariwal2021diffusion}, with a continuous value in place of the label: the
additive case of feature-wise linear modulation~\citep{perez2018film}. $W_2$ and $b_2$ are
zero-initialized, so at the start of training $\psi\equiv0$ and Eq.~\eqref{eq:imgbranchcsi} reduces
to Eq.~\eqref{eq:imgbranch}.

Adding this path to the two ends of Eq.~\eqref{eq:axis} gives the cross-modal variants of
Table~\ref{tab:designspace}: \ArmFiLM{} ($\vect{v}=\mathcal{T}(\hat{\vect{x}}_0)$, trained twice) and
\ArmFiLMPast{} ($\vect{v}=\vect{a}'$). Because $\hat{\vect{y}}_0$ is detached, the path carries forward
values only and closes no gradient cycle. Figure~\ref{Fig:crossmodal} shows one step of \ArmFiLM{}.

\paragraph{The cross-modal comparisons.} Table~\ref{tab:pairs} sets out the three comparisons of
Table~\ref{tab:designspace}. Each input choice of Eq.~\eqref{eq:axis} runs its own phase~1, whereas the
cross-modal variants start phase~2 from one shared phase-1 checkpoint, that of \propPast{} at seed~1.
\ArmFiLMmatched{} is a second training of the variant from that checkpoint, with the budget and
selection rule of \propFuture{}; besides the path, it differs from \propFuture{} only in its starting
checkpoint. \ArmFiLM{} is the first training, run to a longer budget, resumed twice and selected within
the same $960$-epoch ceiling. At matched training budget the path costs $4.5$~W\,m$^{-2}$ in RMSE, with an
interval that excludes zero, more than the noise floors of Section~\ref{sec:noise}, and no difference
in CRPSS. The first training and the past-sky pair differ from their partners by no more than the
largest of those floors. In \ArmFiLMPast{} the fed-back estimate changes only the image branch, which
that arm's irradiance head does not read, so any difference in its forecast comes from training, not
from the path at sampling time.

\begin{figure}[pos=!htbp]
\centering
\begin{tikzpicture}[
  font=\footnotesize, >={Stealth[length=2mm]}, node distance=4mm,
  box/.style={draw, rounded corners=2pt, align=center, inner sep=3pt, minimum height=8mm},
  add/.style={box, draw=orange!85!black, thick, fill=orange!6},
  lab/.style={align=center, inner sep=1pt},
  arr/.style={->},
  addarr/.style={->, draw=orange!85!black, thick, dashed}]
\node[lab] (inA) at (0,2.6) {$\vect{y}_t,\ \vect{e},\ \vect{a}'$};
\node[box, minimum width=31mm] (H1) at (3.3,2.6) {$H_\phi$, conditioning pass\\Eq.~\eqref{eq:condpass}};
\node[add, minimum width=30mm] (Y0) at (8.6,2.6) {$\hat{\vect{y}}_0$ from $\vect{y}_t$ and $\mathrm{sg}[\epsilon'_y]$\\Eq.~\eqref{eq:y0hat}};
\node[box] (gam) at (3.9,1.0) {$\gamma(t)$};
\node[add] (psi) at (8.6,1.0) {$\psi(\hat{y}^j_0)$};
\node[draw=orange!85!black, thick, circle, inner sep=1pt] (plus) at (5.6,1.0) {$+$};
\node[lab] (xt) at (0,-0.6) {$\vect{x}_t$};
\node[box, minimum width=31mm] (U) at (3.3,-0.6) {UNet $+$ ControlNet\\(frame $j$)};
\node[box, minimum width=22mm] (R) at (7.3,-0.6) {temporal refiner};
\node[lab] (ctx) at (1.7,-1.8) {context $[\,\vect{e},s^j_t\,]$};
\node[lab] (ctl) at (4.6,-1.8) {control image $\vect{a}^i$};
\node[box, minimum width=30mm] (X0) at (10.9,-0.6) {$\hat{\vect{x}}_0$ from $\vect{x}_t$ and $\mathrm{sg}[\epsilon_x]$\\Eq.~\eqref{eq:x0hatimg}};
\node[box, minimum width=31mm] (H2) at (10.9,-2.6) {$H_\phi$, prediction pass\\reads $\mathcal{T}(\hat{\vect{x}}_0)$, Eq.~\eqref{eq:predpass}};
\node[lab] (inD) at (7.3,-2.6) {$\vect{y}_t,\ \vect{e}$};
\node[lab] (ey) at (13.6,-2.6) {$\epsilon_y$};
\node[draw, dashed, rounded corners=3pt, inner sep=5pt, fit=(gam)(psi)(plus)(U)(R)(ctx)(ctl)] (G) {};
\node[anchor=north west, inner sep=2pt] at (G.north west) {$G_\theta$, Eq.~\eqref{eq:imgbranchcsi}};
\draw[arr] (inA) -- (H1);
\draw[arr] ([xshift=-8mm]H1.south) -- node[lab, left] {$s_t$} ([xshift=-8mm]U.north);
\draw[arr] (gam) -- (plus);
\draw[arr] (xt) -- (U);
\draw[arr] (ctx.north) -- (ctx.north |- U.south);
\draw[arr] (ctl.north) -- (ctl.north |- U.south);
\draw[arr] (U) -- (R);
\draw[arr] (R) -- node[lab, above] {$\epsilon_x$} (X0);
\draw[arr] (X0) -- (H2);
\draw[arr] (inD) -- (H2);
\draw[arr] (H2) -- (ey);
\draw[addarr] (H1) -- node[lab, above] {$\epsilon'_y$} (Y0);
\draw[addarr] (Y0) -- node[lab, right] {$\hat{y}^j_0$} (psi);
\draw[addarr] (psi) -- (plus);
\draw[addarr] (plus.south) -- ++(0,-0.45) -| node[lab, pos=0.25, below] {$\tilde{\gamma}^j(t)$, Eq.~\eqref{eq:csiemb}} ([xshift=10mm]U.north);
\end{tikzpicture}
\caption{One denoising step of \ArmFiLM{}. Black: the step of the proposed model,
Eqs.~\eqref{eq:condpass}--\eqref{eq:predpass}. Orange, dashed: what the variant adds. The
conditioning pass's noise estimate $\epsilon'_y$, detached, gives $\hat{\vect{y}}_0$
(Eq.~\eqref{eq:y0hat}); inside $G_\theta$, $\psi(\hat{y}^j_0)$ is added to the timestep embedding
$\gamma(t)$, and the UNet call for frame $j$ reads the sum $\tilde{\gamma}^j(t)$
(Eq.~\eqref{eq:csiemb}). Both links between the irradiance head and the image branch pass through a stop-gradient, so no
gradient cycle closes. In \ArmFiLMPast{} the prediction pass is absent: one pass reading
$\vect{a}'$ supplies $s_t$, $\epsilon_y$ and, through Eq.~\eqref{eq:y0hat}, $\hat{\vect{y}}_0$.}
\label{Fig:crossmodal}
\end{figure}

\begin{table}[h]
\centering
\caption{The three cross-modal comparisons, on the matched test subset. $\Delta$ is with-path minus
without-path, with its paired day-block interval. The last column lists what differs between the two
arms \emph{besides} the path.}
\label{tab:pairs}
\footnotesize
\setlength{\tabcolsep}{2pt}
\begin{tabular}{@{}Q{0.17\linewidth}Q{0.25\linewidth}P{0.17\linewidth}P{0.21\linewidth}Q{0.16\linewidth}@{}}
\toprule
without path & with path & $\Delta$RMSE\,\arrowzero{} & $\Delta$CRPSS\,\arrowzero{} & other differences \\
\midrule
\DTLforeach*{prs}{\Wo=without,\Wi=withp,\Dr=drmse,\Dc=dcrpss,\Cf=confounds}{%
  \DTLiffirstrow{}{\\}\csname\Wo\endcsname & \csname\Wi\endcsname & \Dr & \Dc & \Cf}\\
\bottomrule
\end{tabular}
\end{table}

No comparison isolates the path completely: the arms in each pair also differ in training history,
and each variant exists at one training seed. The path brings no forecast gain in any of the three and
costs point accuracy in the matched one, although one seed does not establish the size of that cost.
We make no claim about its effect on image quality.

\section{Protocol details}
\label{sec:app-protocol}

\begin{table}[t]
\centering
\caption{The five training stages: three image-branch stages, then joint diffusion training in two
phases. Each stage starts from the checkpoint the previous one selected.
Validation criteria are computed on the validation split (Section~\ref{sec:setup}).}
\label{tab:training}
\footnotesize
\setlength{\tabcolsep}{2pt}
\begin{tabular}{@{}Q{\wTrSt}Q{\wTrTr}Q{\wTrFr}Q{\wTrLr}Q{\wTrSp}@{}}
\toprule
Stage & Trains & Frozen & Learning rate & Stopping and checkpoint \\
\midrule
\DTLforeach*{trn}{\St=stage,\Tr=trains,\Fr=frozen,\Lr=lr,\Sp=stop}{%
  \DTLiffirstrow{}{\\}\St & \Tr & \Fr & \Lr & \Sp}\\
\bottomrule
\end{tabular}
\end{table}

Each training step co-noises the target latents and the clear-sky-index sequence at one shared
timestep and minimizes the loss of Eq.~\eqref{eq:loss}. In both phases the validation criterion is the
LPIPS~\citep{zhang2018lpips} of the generated frames plus the RMSE of the forecast clear-sky index.
Every arm starts from the same ControlNet checkpoint with the same ceiling of $50$ epochs and early
stopping on that criterion, so no arm of Sections~\ref{sec:comparison}--\ref{sec:ablations} trains
longer than another. Every method in this paper, ours and the baselines alike, is scored at its own best-validation
checkpoint, with one exception: the irradiance-stream arm and its control (Section~\ref{sec:inputs}),
described below. For the baselines this is the last epoch whose validation loss improved by
more than $10^{-4}$; for \ConvLSTM{} it is epoch $16$, $1.3\%$ above its lowest validation loss at
epoch $22$. \propFuture{} is scored at epoch $959$ and \propPast{} at $954$, at all three seeds; at
seed~1 the latter's second phase stopped five epochs short of the ceiling. The arms of Appendix~\ref{sec:app-designspace} follow the same stages, with
differences in budget or starting checkpoint stated in Table~\ref{tab:pairs}; \ArmScalar{} has no
image branch and trains in a single phase.

\paragraph{Sampler.} All joint arms sample at $50$ reverse diffusion steps, the point beyond which a
sweep over step counts showed no material improvement in forecast metrics; the setting is fixed for
every arm and seed.

\paragraph{Ensemble size.} Probabilistic forecasts use $M=10$ members, each from an independent
initial noise. CRPS uses the fair (bias-corrected) estimator; the naive estimator gives materially
more optimistic skill at this ensemble size.

\begin{table}[h]
\centering
\caption{Per-horizon values behind statements made as ranges in the main text: the frozen U-Net
regressor applied to the ground-truth and to the generated future frames (RMSE, W\,m$^{-2}$, of the
mean of the ten per-member regressions for the generated frames; Section~\ref{sec:cascade}); the CRPSS of \propFuture{} (Section~\ref{sec:horizon}); its empirical coverage at the nominal
$90\%$ level, $\mathrm{PICP}_{90}$ (Section~\ref{sec:reliability}); and the cross-year minus in-period
forecast skill of \propFuture{} and \SunSet{} (Section~\ref{sec:crossyear}). Matched test subset,
$50$ sampling steps.}
\label{tab:perhorizon}
\footnotesize
\setlength{\tabcolsep}{2pt}
\begin{tabular}{@{}Q{\wPhzH}P{\wPhzN}P{\wPhzN}P{\wPhzN}P{\wPhzN}P{\wPhzN}P{\wPhzN}@{}}
\toprule
& \multicolumn{2}{c}{frozen regressor on} & & & \multicolumn{2}{c}{$\Delta$FS, cross-year} \\
\cmidrule(lr){2-3}\cmidrule(lr){6-7}
Horizon & true frames & generated & CRPSS & $\mathrm{PICP}_{90}$ & \propFuture{} & \SunSet{} \\
\midrule
\DTLforeach*{phz}{\Hz=h,\Rg=rgt,\Rn=rgen,\Cs=crpss,\Pb=p90,\Df=dfs,\Ds=dfssun}{%
  \DTLiffirstrow{}{\\}\Hz & \Rg & \Rn & \Cs & \Pb & \Df & \Ds}\\
\bottomrule
\end{tabular}
\end{table}

\begin{table}[h]
\centering
\caption{Point estimates on the matched test subset with all $1998$ windows, without the $13$ that are
not contiguous (contiguous, $N=1985$), and without the $62$ whose targets or persistence origin include
a minute SIRTA has since marked unusable (QC-valid, $N=1936$); Section~\ref{sec:data}. RMSE in
W\,m$^{-2}$; CRPSS against smart persistence on the same windows. \underline{\textbf{Bold underlined}}: best in each column; \textbf{bold}: second best (smart persistence is not
ranked). No method changes rank on either
score in either comparison, and none of the $13$ non-contiguous windows is in the ramp-event subset of
Section~\ref{sec:ramps}.}
\label{tab:contiguity}
\footnotesize
\setlength{\tabcolsep}{2pt}
\begin{tabular}{@{}Q{\wCtgM}P{\wCtgN}P{\wCtgN}P{\wCtgN}P{\wCtgN}P{\wCtgN}P{\wCtgN}@{}}
\toprule
& \multicolumn{3}{c}{RMSE\,\arrowdown{}} & \multicolumn{3}{c}{CRPSS\,\arrowup{}} \\
\cmidrule(lr){2-4}\cmidrule(lr){5-7}
Method & all & contiguous & QC-valid & all & contiguous & QC-valid \\
\midrule
\DTLforeach*[\equal{\Grp}{0}]{ctg}{\Grp=group,\Mth=method,\Ra=rall,\Rx=rex,\Ca=call,\Cx=cex,\Rq=rqc,\Cq=cqc,\BRa=brall,\BRx=brex,\BRq=brqc,\BCa=bcall,\BCx=bcex,\BCq=bcqc}{%
  \DTLiffirstrow{}{\\}\csname\Mth\endcsname & \csname\BRa\endcsname{\Ra} & \csname\BRx\endcsname{\Rx} & \csname\BRq\endcsname{\Rq} & \csname\BCa\endcsname{\Ca} & \csname\BCx\endcsname{\Cx} & \csname\BCq\endcsname{\Cq}}\\
\midrule
\DTLforeach*[\equal{\Grp}{1}]{ctg}{\Grp=group,\Mth=method,\Ra=rall,\Rx=rex,\Ca=call,\Cx=cex,\Rq=rqc,\Cq=cqc,\BRa=brall,\BRx=brex,\BRq=brqc,\BCa=bcall,\BCx=bcex,\BCq=bcqc}{%
  \DTLiffirstrow{}{\\}\csname\Mth\endcsname & \csname\BRa\endcsname{\Ra} & \csname\BRx\endcsname{\Rx} & \csname\BRq\endcsname{\Rq} & \csname\BCa\endcsname{\Ca} & \csname\BCx\endcsname{\Cx} & \csname\BCq\endcsname{\Cq}}\\
\midrule
\DTLforeach*[\equal{\Grp}{2}]{ctg}{\Grp=group,\Mth=method,\Ra=rall,\Rx=rex,\Ca=call,\Cx=cex,\Rq=rqc,\Cq=cqc,\BRa=brall,\BRx=brex,\BRq=brqc,\BCa=bcall,\BCx=bcex,\BCq=bcqc}{%
  \DTLiffirstrow{}{\\}\csname\Mth\endcsname & \csname\BRa\endcsname{\Ra} & \csname\BRx\endcsname{\Rx} & \csname\BRq\endcsname{\Rq} & \csname\BCa\endcsname{\Ca} & \csname\BCx\endcsname{\Cx} & \csname\BCq\endcsname{\Cq}}\\
\midrule
\DTLforeach*[\equal{\Grp}{3}]{ctg}{\Grp=group,\Mth=method,\Ra=rall,\Rx=rex,\Ca=call,\Cx=cex,\Rq=rqc,\Cq=cqc,\BRa=brall,\BRx=brex,\BRq=brqc,\BCa=bcall,\BCx=bcex,\BCq=bcqc}{%
  \DTLiffirstrow{}{\\}\csname\Mth\endcsname & \csname\BRa\endcsname{\Ra} & \csname\BRx\endcsname{\Rx} & \csname\BRq\endcsname{\Rq} & \csname\BCa\endcsname{\Ca} & \csname\BCx\endcsname{\Cx} & \csname\BCq\endcsname{\Cq}}\\
\midrule
\DTLforeach*[\equal{\Grp}{4}]{ctg}{\Grp=group,\Mth=method,\Ra=rall,\Rx=rex,\Ca=call,\Cx=cex,\Rq=rqc,\Cq=cqc,\BRa=brall,\BRx=brex,\BRq=brqc,\BCa=bcall,\BCx=bcex,\BCq=bcqc}{%
  \DTLiffirstrow{}{\\}\csname\Mth\endcsname & \csname\BRa\endcsname{\Ra} & \csname\BRx\endcsname{\Rx} & \csname\BRq\endcsname{\Rq} & \csname\BCa\endcsname{\Ca} & \csname\BCx\endcsname{\Cx} & \csname\BCq\endcsname{\Cq}}\\
\midrule
\DTLforeach*[\equal{\Grp}{5}]{ctg}{\Grp=group,\Mth=method,\Ra=rall,\Rx=rex,\Ca=call,\Cx=cex,\Rq=rqc,\Cq=cqc,\BRa=brall,\BRx=brex,\BRq=brqc,\BCa=bcall,\BCx=bcex,\BCq=bcqc}{%
  \DTLiffirstrow{}{\\}\csname\Mth\endcsname & \csname\BRa\endcsname{\Ra} & \csname\BRx\endcsname{\Rx} & \csname\BRq\endcsname{\Rq} & \csname\BCa\endcsname{\Ca} & \csname\BCx\endcsname{\Cx} & \csname\BCq\endcsname{\Cq}}\\
\midrule
\DTLforeach*[\equal{\Grp}{6}]{ctg}{\Grp=group,\Mth=method,\Ra=rall,\Rx=rex,\Ca=call,\Cx=cex,\Rq=rqc,\Cq=cqc,\BRa=brall,\BRx=brex,\BRq=brqc,\BCa=bcall,\BCx=bcex,\BCq=bcqc}{%
  \DTLiffirstrow{}{\\}\csname\Mth\endcsname & \csname\BRa\endcsname{\Ra} & \csname\BRx\endcsname{\Rx} & \csname\BRq\endcsname{\Rq} & \csname\BCa\endcsname{\Ca} & \csname\BCx\endcsname{\Cx} & \csname\BCq\endcsname{\Cq}}\\
\bottomrule
\end{tabular}
\end{table}

\paragraph{Solar elevation.} Table~\ref{tab:elevation} repeats the main scores after removing the
windows whose solar elevation at the forecast origin, computed with pvlib for the SIRTA site, is below
$10^\circ$ ($N=1639$) or $20^\circ$ ($N=1154$, $32$ test days). Absolute errors grow with the
elevation floor because the remaining windows carry more irradiance, but the skill scores and every
paired comparison keep their sign. \propFuture{} minus \SunSet{} is $+6.5$, $+7.1$ and
$+8.4$~W\,m$^{-2}$ in RMSE and $+0.281$, $+0.285$ and $+0.282$ in CRPSS on the three sets;
\propFuture{} minus \PhyDNetGAN{} is $-8.3$, $-8.8$ and $-9.6$~W\,m$^{-2}$, all with day-block
intervals excluding zero. The tertile CRPSS lead of \propFuture{} over \SunSet{} stays between $0.25$
and $0.32$, and all $334$ ramp windows have an origin elevation of at least $10^\circ$.

\begin{table}[h]
\centering
\caption{Main scores on the matched test subset with all $1998$ windows and with the windows whose
solar elevation at the forecast origin is below $10^\circ$ or $20^\circ$ removed. RMSE in W\,m$^{-2}$;
CRPSS against smart persistence on the same windows. All methods of Table~\ref{tab:comparison} were
scored; the rows shown are the reference and the main comparators, and the order of the others changes
only as stated in Section~\ref{sec:protocol}. \underline{\textbf{Bold underlined}}: best in each column; \textbf{bold}: second best among the rows shown (smart persistence is not ranked).}
\label{tab:elevation}
\footnotesize
\setlength{\tabcolsep}{2pt}
\begin{tabular}{@{}Q{\wElvM}P{\wElvN}P{\wElvN}P{\wElvN}P{\wElvN}P{\wElvN}P{\wElvN}@{}}
\toprule
& \multicolumn{2}{c}{all ($N=1998$)} & \multicolumn{2}{c}{$\ge10^\circ$ ($N=1639$)} & \multicolumn{2}{c}{$\ge20^\circ$ ($N=1154$)} \\
\cmidrule(lr){2-3}\cmidrule(lr){4-5}\cmidrule(lr){6-7}
Method & RMSE & CRPSS & RMSE & CRPSS & RMSE & CRPSS \\
\midrule
\DTLforeach*{elv}{\Arm=arm,\Ra=r0,\Ca=c0,\Rb=r10,\Cb=c10,\Rc=r20,\Cc=c20,\BRa=br0,\BCa=bc0,\BRb=br10,\BCb=bc10,\BRc=br20,\BCc=bc20}{%
  \DTLiffirstrow{}{\\}\csname\Arm\endcsname & \csname\BRa\endcsname{\Ra} & \csname\BCa\endcsname{\Ca} & \csname\BRb\endcsname{\Rb} & \csname\BCb\endcsname{\Cb} & \csname\BRc\endcsname{\Rc} & \csname\BCc\endcsname{\Cc}}\\
\bottomrule
\end{tabular}
\end{table}

\paragraph{Irradiance-stream arm and its control.} The arm of Section~\ref{sec:inputs} adds a third
token stream to the irradiance head: the eight past values of measured irradiance, encoded as the
clear-sky index and the raw irradiance. They are mapped to eight tokens with their own learned source
embedding, concatenated with the tokens of the denoised estimate, and also added on the query side so
that every forecast step receives them. All other settings are unchanged. Because the stream needs a
modified code path, each seed is compared with a control trained on the same path with the stream
disabled. The control still constructs the unused modules, so both arms start from identical weights
and consume identical random draws; from the same weights, the modified and original samplers produce
bit-identical ensembles.

Relative to the original code, the stream's modules are built before the irradiance head, the temporal
refiner and the source embedding, so at the same seed those three modules receive different random
initial weights. Training noise, timesteps and data order come from separately seeded generators and
are unchanged. As a result the control lies above the headline arm at each seed: $108.51$ against
$104.58$~W\,m$^{-2}$ at seed~1, and $108.13$ against $106.53$~W\,m$^{-2}$ at seed~2, under the same
sampler seeding. Copying the headline arm's initial weights into the modified path, and otherwise
training and scoring unchanged, gives $104.63$~W\,m$^{-2}$ at seed~1: within $0.05$ of the headline arm
and $3.88$ below the control. The offset is therefore an initialization difference at that seed, not a
property of the code path or of the ablation; the arm was not retrained at seed~2. The comparison of
Section~\ref{sec:inputs} is paired within the code path and is unaffected either way. Both arms of each pair are scored at the ceiling epoch, $959$, the epoch of the headline arm, so that
the pair is compared at the same point in training. The control's best-validation epochs are $954$ at
seed~1 and $948$ at seed~2; scored at epoch $954$, the seed-1 control reaches $107.72$~W\,m$^{-2}$,
$0.78$ below its ceiling-epoch score. Re-running the sampler from a different random state moves RMSE by
$1.34$~W\,m$^{-2}$ (Section~\ref{sec:noise}).

\paragraph{Compute.} The proposed generator is the largest model in the comparison
($198{,}834{,}269$ parameters) and not the fastest at inference. The single-stage design is a
structural claim about error propagation, not a compute claim: the regressor a cascade requires is a
negligible fraction of total cost.

\section{The information each method conditions on}
\label{sec:lattice}

Several comparisons in this paper are between models whose \emph{information sets} differ, sometimes
more than their architectures do. Table~\ref{tab:lattice} states those sets so that a difference in
score can be attributed to the right cause.

Symbols: $\vect{a}$ the eight past sky frames; $\vect{b}$ the measured history of the forecast
variable; $\vect{e}$ the deterministic covariates at the forecast origin (clock variables and the
clear-sky irradiance); $\vect{x}_t,\vect{y}_t$ the noised future frames and future clear-sky index at diffusion
step $t$; $\hat{\vect{x}}_0,\hat{\vect{y}}_0$ the model's own denoised estimates of those; $\vect{x}_\tau,\vect{y}_\tau$
\PVMM{}'s noised sky-image and PV-power iterates at its own diffusion step $\tau$; $s_t$ the CSI
tokens and $\vect{a}^i$ the control image (Section~\ref{sec:method-step}); ``$\vect{a}$ ind.'' marks
past frames that reach the irradiance head only indirectly, through $\hat{\vect{x}}_0$. ``Loop'' is gradient
topology: \emph{closed} if a gradient cycle runs between the two branches, \emph{open} if at least
one cross-link carries forward values only.

\begin{table*}[t]
\centering
\caption{What each method conditions on. Rows above the rule are arms of this work; below are the
external comparators. Our arms receive no $\vect{b}$: the proposed model never sees a measured
irradiance value. Section~\ref{sec:inputs} isolates what that costs.}
\label{tab:lattice}
\footnotesize
\setlength{\tabcolsep}{2pt}
\begin{tabular}{@{}Q{\wLatArm}Q{\wLatSc}Q{\wLatIm}P{\wLatL}P{\wLatB}@{}}
\toprule
Arm & Irradiance head reads & Image branch reads & loop & $\vect{b}$? \\
\midrule
\DTLforeach*[\equal{\Grp}{1}]{lat}{\Grp=group,\Arm=arm,\Sc=scalar,\Im=image,\Lp=loop,\Bb=bflag}{%
  \DTLiffirstrow{}{\\}\csname\Arm\endcsname & \Sc & \Im & \Lp & \Bb}\\
\midrule
\DTLforeach*[\equal{\Grp}{2}]{lat}{\Grp=group,\Arm=arm,\Sc=scalar,\Im=image,\Lp=loop,\Bb=bflag}{%
  \DTLiffirstrow{}{\\}\csname\Arm\endcsname & \Sc & \Im & \Lp & \Bb}\\
\bottomrule
\end{tabular}
\end{table*}

Two entries deserve a note. \PVMM{}'s coupled reverse step conditions each branch on the historical
series and on the \emph{noised iterate} of both modalities, and forms no denoised estimate; among our
arms, \ArmConcat{} has the matching sky-image information set. The two cross-modal variants feed back
$\hat{\vect{y}}_0$, an \emph{irradiance} quantity rather than an image one.

\section{Architecture details}
\label{sec:app-arch}

The image tokenizer $\mathcal{T}$ maps each $4\times8\times8$ latent frame to sixteen tokens of width
$256$. The irradiance head $H_\phi$ is a four-block transformer of width $256$ over the eight
clear-sky-index positions. It self-attends across the positions and cross-attends to a sequence of
image tokens, with a learned two-entry embedding marking whether a token describes past or future
frames.

The UNet cross-attends to the two-token context $[\,\vect{e},\,s^j_t\,]\in\mathbb{R}^{2\times4}$,
which it projects to its own attention width. The ControlNet takes the control image $\vect{a}^i$ as
its spatial conditioning input, also attends to $\vect{e}$, and reaches the UNet as additive down- and
mid-block residuals. At each of the $8\times8$ latent positions, the temporal refiner concatenates every
frame's noisy latent and per-frame estimate, $[\,\vect{x}^j_t,\,\tilde{\epsilon}^j\,]\in\mathbb{R}^{8}$,
and projects them to width $128$. It then adds a learned frame-position embedding and applies two
self-attention blocks across the eight frames, across time only, not across space. Its output layer is
zero-initialized, so it starts as the identity, $\epsilon_x=\tilde{\epsilon}$, and learns only a
cross-frame correction.